\documentclass[letterpaper]{article} % DO NOT CHANGE THIS
\usepackage{aaai2027}  % DO NOT CHANGE THIS
\usepackage[hyphens]{url}  % DO NOT CHANGE THIS
\usepackage{graphicx} % DO NOT CHANGE THIS										% 插入外部图片所需
\usepackage{natbib}  % DO NOT CHANGE THIS AND DO NOT ADD ANY OPTIONS TO IT
\usepackage{caption} % DO NOT CHANGE THIS AND DO NOT ADD ANY OPTIONS TO IT
\usepackage{algorithm}
\usepackage{algorithmic}

\usepackage{newfloat}
\usepackage{listings}
\DeclareCaptionStyle{ruled}{labelfont=normalfont,labelsep=colon,strut=off} % DO NOT CHANGE THIS
\floatstyle{ruled}
\newfloat{listing}{tb}{lst}{}
\floatname{listing}{Listing}

\usepackage{booktabs}

\usepackage{xcolor}		% 2. 颜色定义所需 (HTML 模式)
\usepackage{tikz}		% 3. 绘图核心宏包
\usepackage{pgfplots}	% 3. 绘图核心宏包
\usetikzlibrary{arrows.meta} % 用于绘制带有 -Stealth 样式的箭头
\pgfplotsset{compat=1.18} % 版本号可以根据你本地的 TeX Live/MikTeX 版本向下兼容，如 1.17, 1.16 等
\usepackage{amsmath}
\usepackage{amssymb}
\usepackage{multirow}
\usepackage{array}
\usepackage[table]{xcolor}
\newlength{\HeaderShift}   %设置表格划线距离
\usepackage{pifont}

\title{SPEANet: Structural Prior Enhanced Attention Network for Parameter-Efficient Remote Sensing Object Detection}

\author{
	Wei Lu,
	Junjie Li,
	Feifei Sang,
	Si-Bao Chen\textsuperscript{*}
}
\affiliations{MOE Key Lab of ICSP, IMIS Lab of Anhui, Anhui Provincial Key Lab of Multimodal Cognitive Computation, Zenmorn-AHU AI Joint Lab, School of Computer Science and Technology, Anhui University, Hefei 230601, China\\
	luwei@ahu.edu.cn, e125221143@stu.ahu.edu.cn, ffeisang@163.com, sbchen@ahu.edu.cn
}

\begin{document}

\maketitle

\begin{abstract}
	Remote sensing object detection (RSOD) requires compact backbones capable of preserving weak geometric cues under extreme scale variation and background clutter. Fixed structural operators provide complementary contour and frequency responses without introducing learnable operator coefficients. However, directly injecting these responses can amplify content-irrelevant textures, while applying a uniform operator design across the hierarchy may be poorly matched to stage-specific representation requirements. We propose the Structural Prior Enhanced Attention Network (SPEANet), a parameter-efficient RSOD backbone that integrates fixed operators through stage-specific prior extraction and context-conditioned response modulation. SPEANet assigns smoothed contour and multi-order directional modeling to shallow, high-resolution features, while employing a compact approximation-detail interaction mechanism in deeper stages. Learned spatial gates regulate the resulting prior responses before residual fusion. Experiments on five benchmarks, together with evaluations across seven detection frameworks on DOTA-v1.0, achieve a favorable accuracy-parameter trade-off. With Oriented R-CNN, SPEANet achieves 78.55\% mAP on DOTA-v1.0, 72.24\% mAP on DOTA-v1.5, and 67.30\% mAP on DIOR-R using 23.0M total parameters, including a 5.97M-parameter backbone.
\end{abstract}

\begin{links}
	\link{Code}{https://github.com/AeroVILab-AHU/SPEANet}
\end{links}

\begin{figure}[t]
	\centering
	\makebox[\linewidth][c]{%
		\begin{tikzpicture}[
			basept/.style={circle, draw=black!55, line width=0.45pt, minimum size=5.8pt, inner sep=0pt},
			ourspt/.style={circle, draw=speanetRed!80!black, line width=1.0pt, minimum size=7.6pt, inner sep=0pt},
			baselab/.style={font=\scriptsize, fill=white, fill opacity=0.86, text opacity=1, inner sep=1pt},
			ourslab/.style={font=\scriptsize\bfseries, fill=speanetRed!8, draw=speanetRed!50, rounded corners=1pt, text=speanetRed!70!black, inner sep=1.4pt}
			]
			\definecolor{speanetRed}{HTML}{E76F51}
			\definecolor{baseBlue}{HTML}{4E79A7}
			\definecolor{baseGreen}{HTML}{59A14F}
			\definecolor{basePurple}{HTML}{8A60B8}
			\definecolor{baseGold}{HTML}{B98B19}
			\begin{axis}[
				width=0.98\linewidth,
				height=4.85cm,
				xmin=20, xmax=44, ymin=73.6, ymax=79.1,
				xlabel={Total Params (M)},
				ylabel={mAP (\%)},
				xtick={20,25,30,35,40},
				ytick={74,75,76,77,78,79},
				minor tick num=1,
				grid=both,
				major grid style={line width=0.25pt, draw=gray!24},
				minor grid style={line width=0.15pt, draw=gray!12},
				axis lines=left,
				axis line style={-Stealth, black!75, line width=0.55pt},
				tick style={black!70, line width=0.45pt},
				label style={font=\footnotesize\bfseries},
				tick label style={font=\scriptsize},
				xlabel style={yshift=1pt},
				ylabel style={yshift=-1pt},
				clip=false,
				set layers,
				axis on top=false
				]
				\draw[-Stealth, line width=0.55pt, draw=black!42] (axis cs:35.0,74.45) -- (axis cs:25.0,78.08)
				node[pos=0.55, above, sloped, font=\scriptsize\bfseries, text=black!55, fill=white, fill opacity=0.82, text opacity=1, inner sep=0.8pt] {better trade-off};
				
				\node[basept, fill=baseBlue!62] (s2anet) at (axis cs:38.5,74.12) {};
				\node[baselab, anchor=west, text=baseBlue!75!black] at (axis cs:38.85,74.12) {S$^2$ANet 74.12};
				
				\node[basept, fill=baseBlue!45] (orcnn) at (axis cs:41.1,75.87) {};
				\node[baselab, anchor=west, text=baseBlue!75!black] at (axis cs:41.45,75.87) {O-RCNN 75.87};
				
				\node[basept, fill=baseGreen!70] (redet) at (axis cs:31.6,76.25) {};
				\node[baselab, anchor=west, text=baseGreen!65!black] at (axis cs:31.95,76.25) {ReDet 76.25};
				
				\node[basept, fill=basePurple!72] (lsknet) at (axis cs:31.0,77.49) {};
				\node[baselab, anchor=west, text=basePurple!75!black] at (axis cs:31.35,77.49) {LSKNet-S 77.49};
				
				\node[basept, fill=baseGold!78] (pkinet) at (axis cs:30.8,78.39) {};
				\node[baselab, anchor=south west, text=baseGold!70!black] at (axis cs:31.15,78.37) {PKINet-S 78.39};
				
				\node[ourspt, fill=speanetRed] (speanet) at (axis cs:23.0,78.55) {};
				\draw[densely dashed, draw=speanetRed!70, line width=0.55pt] (axis cs:23.0,73.6) -- (speanet);
				\draw[densely dashed, draw=speanetRed!70, line width=0.55pt] (axis cs:20.0,78.55) -- (speanet);
				\node[ourslab, anchor=south west] at (axis cs:23.35,78.52) {SPEANet 78.55};
			\end{axis}
		\end{tikzpicture}
	}
	\includegraphics[width=1\linewidth]{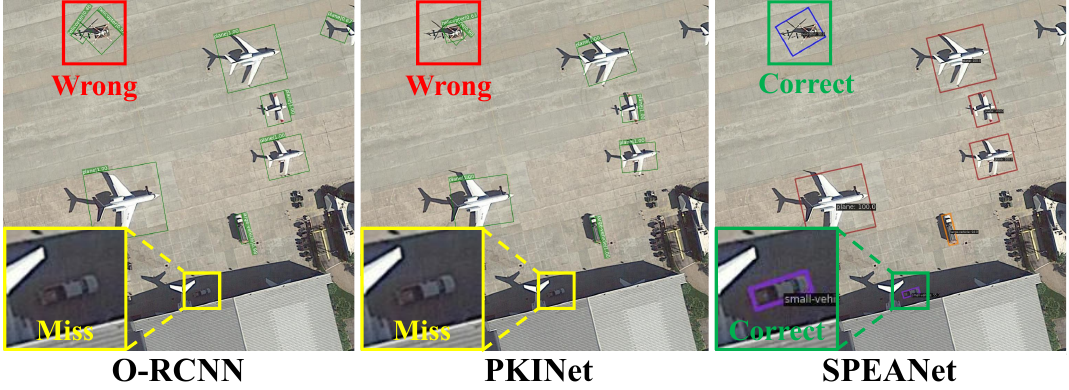}
	\caption{\textbf{Top}: Accuracy-parameter comparison on the DOTA-v1.0 test set~\cite{xia2018dota}. \textbf{Bottom}: Representative detections in scenes containing weak object structures and background clutter. Red, yellow, and green denote false positives, false negatives, and true positives, respectively.} 	\label{fig:combined_result}
\end{figure}
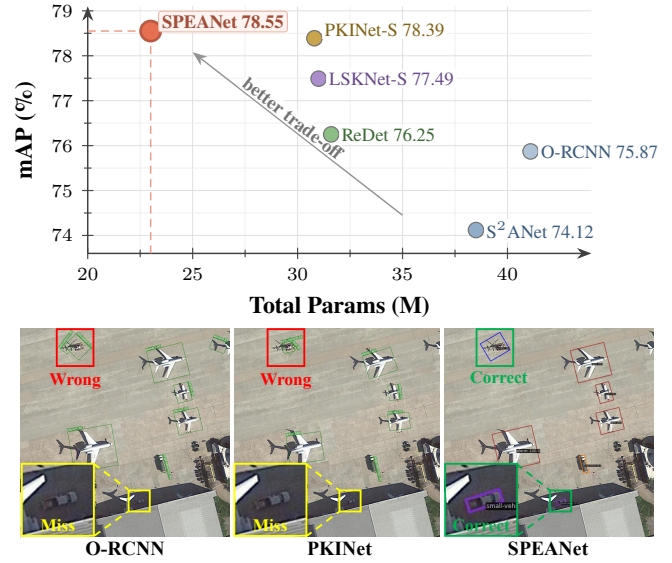

% ---------------------------------Introduction------------------------------
% ---------------------------------Introduction------------------------------
% ---------------------------------Introduction------------------------------
\section{Introduction} \label{sec_Introduction}

Remote sensing object detection (RSOD) aims to localize objects in aerial and satellite imagery. Compared with natural-image detection, RSOD is challenged by arbitrary orientations, extreme scale variation, weak appearance, blur, shadows, low illumination, and dense background clutter. Consequently, discriminative cues are often weak and spatially sparse, particularly for small vehicles, ships, bridges, and densely distributed instances. Resource-constrained UAV and satellite platforms further motivate backbones with controlled parameter counts. As illustrated in Figure~\ref{fig:combined_result}, increasing model capacity alone does not eliminate missed detections and false positives in scenes containing weak object boundaries and complex background patterns.

The challenge arises from the limited separability between weak object structures and high-frequency background patterns. Thin boundaries, corners, and orientation cues can be attenuated by early downsampling before they are sufficiently encoded by deeper layers. Conversely, repetitive patterns from roofs, roads, coastlines, and water surfaces may produce strong gradient responses despite being unrelated to target objects. These effects lead to missed detections when object cues are suppressed and false positives when background patterns dominate the representation. Explicit structural priors may complement learned filters under a constrained parameter budget, but their integration must account for content-dependent noise and feature-stage differences.

Fixed image operators provide complementary inductive biases for structural modeling. Laplacian-of-Gaussian (LoG), Sobel, Laplacian, and Gaussian filters encode smoothed contours, directional gradients, second-order variations, and low-pass responses without adding learnable operator coefficients. Prior-guided RSOD backbones have incorporated edge or Gaussian cues into learned representations~\cite{lu2025legnet,LU2026431}. However, fixed responses are content-agnostic: strong activations may originate from either target structures or repetitive background textures. Their usefulness also varies across feature stages with different spatial resolutions and contextual ranges. Effective integration therefore requires both stage-appropriate prior assignment and content-adaptive response control.

Motivated by this requirement, we formulate prior integration as a generate-modulate-fuse process. Fixed operators first generate complementary structural responses. Learned projections align these responses with the backbone features, while context-conditioned gates regulate their spatial contributions. Residual fusion subsequently incorporates the modulated responses without replacing the learned feature stream. This separation preserves the predefined properties of fixed operators while allowing their contributions to adapt to image content and network depth.

We therefore propose the Structural Prior Enhanced Attention Network (SPEANet), a compact RSOD backbone with a shallow-to-deep allocation of structural priors. Context-Guided Laplacian-of-Gaussian Stem (CLoG-Stem) preserves smoothed contour responses before early downsampling. At the first high-resolution stage, Multi-Order Directional Prior with Frequency-Aware Scaling and Encoding (MDP-FASE) models localized first- and second-order directional variations. At Stages II--IV, Wavelet-Inspired Analysis-Synthesis Mixer (WASM) employs a compact approximation-detail interaction to model broader contextual representations without repeatedly expanding the directional-response bank. Across these prior branches, Context-Guided Frequency Filtering (CGFF) modulates operator responses using learned spatial context before residual fusion. This allocation progresses from localized structural modeling at high resolution to compact frequency interaction in deeper stages.

We evaluate SPEANet on DOTA-v1.0 and v1.5~\cite{xia2018dota}, DIOR-R~\cite{cheng2022anchor}, SODA-A~\cite{cheng2023towards}, and MODA~\cite{han2026moda}. Cross-detector experiments on DOTA-v1.0 further cover seven detection frameworks. With Oriented R-CNN~\cite{xie2021oriented}, SPEANet achieves 78.55\% mAP on DOTA-v1.0, 72.24\% mAP on DOTA-v1.5, 67.30\% mAP on DIOR-R, 37.40\% AP$_{50:95}$ on SODA-A, and 43.8\% mAP on MODA using 23.0M total detector parameters. These results show that SPEANet maintains competitive detection accuracy under a compact parameter budget across variations in object scale, scene distribution, detection framework, and input modality.

Our contributions are summarized as follows:
\begin{itemize}
	\item We propose SPEANet, a parameter-efficient RSOD backbone that formulates fixed-prior integration as stage-specific extraction, context-conditioned modulation, and residual fusion.
	\item We develop a hierarchical prior allocation in which CLoG-Stem and MDP-FASE model localized structures in shallow, high-resolution features, while WASM provides compact approximation-detail interaction in deeper stages. CGFF adaptively regulates content-irrelevant operator responses throughout the hierarchy.
	\item Evaluations on five benchmarks, together with cross-detector comparisons across seven detection frameworks on DOTA-v1.0, demonstrate favorable accuracy-parameter trade-offs across different object scales, scene distributions, and input modalities.
\end{itemize}

\begin{figure*}[t]\centering
	\includegraphics[width=1.0\textwidth]{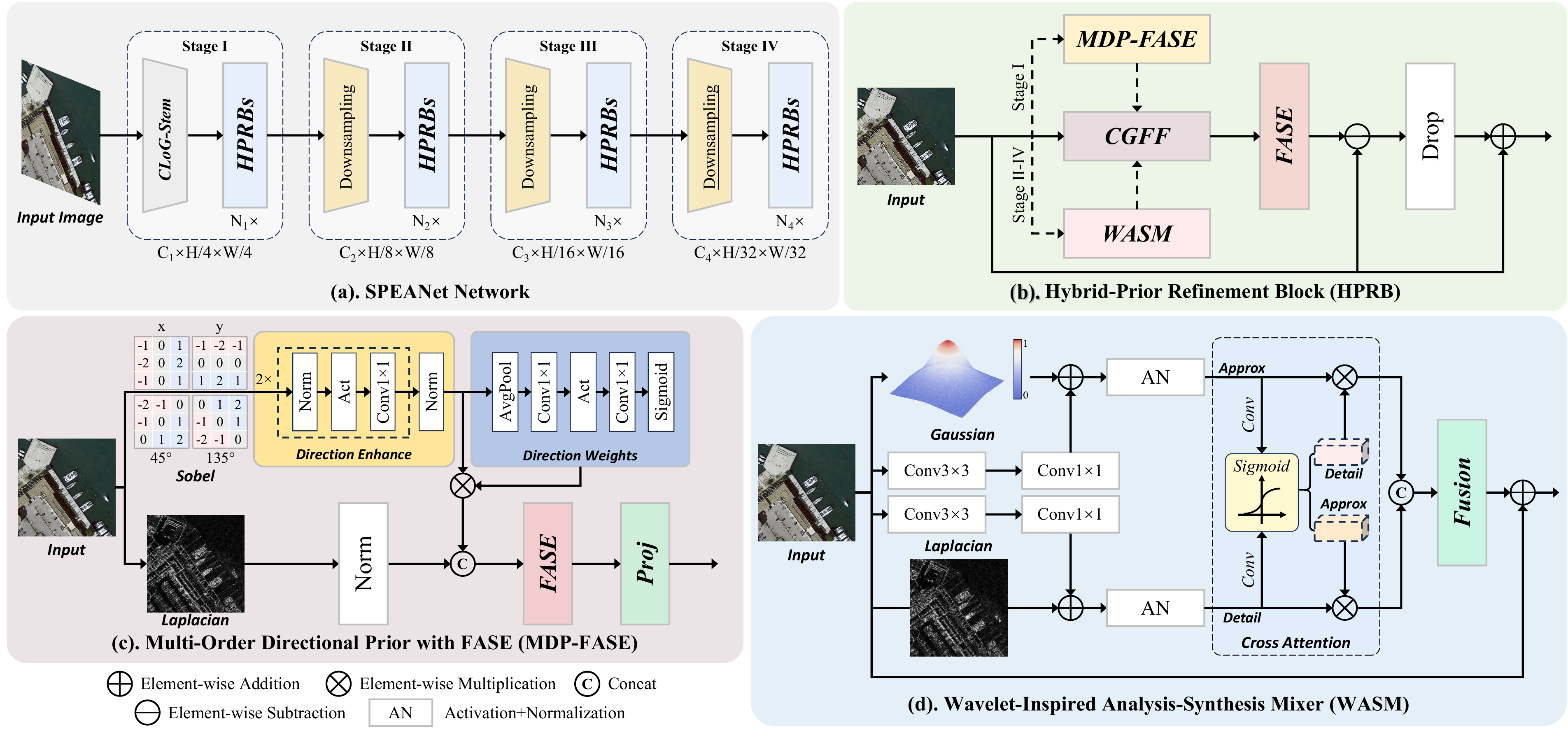}	
	\caption{Overview of SPEANet. (a) Four-stage hierarchical backbone with stage-specific operator assignment. (b) HPRB, which combines prior extraction, context-conditioned modulation, and residual encoding. (c) MDP-FASE for multi-order directional modeling at Stage I. (d) WASM for approximation-detail interaction at Stages II-IV.}\label{fig_overall}
	
	%\caption{Overview of SPEANet. (a) Four-stage prior-guided backbone. (b) Hybrid-Prior Refinement Block (HPRB), which performs stage-wise prior extraction and context-guided reliability filtering. (c) MDP-FASE for high-resolution derivative-prior extraction. (d) WASM for semantic-stage frequency mixing.}	
\end{figure*}

% ---------------------------------Related Work------------------------------
% ---------------------------------Related Work------------------------------
% ---------------------------------Related Work------------------------------
\section{Related Work}   \label{Related_Work}

Oriented object detectors represent objects with rotated boxes to accommodate arbitrary orientations in aerial imagery. Rotated RPN~\cite{yang2018automatic} extends region proposal generation with rotated anchors, while RoI Trans.~\cite{ding2019learning} converts horizontal proposals into rotated RoIs. S$^2$ANet~\cite{han2021align} aligns convolutional features with object orientations, and R$^3$Det~\cite{yang2021r3det} refines rotated detections through feature reconstruction. Oriented R-CNN~\cite{xie2021oriented} directly regresses rotated proposals. Complementary studies address angular periodicity and boundary discontinuity through Gaussian box~\cite{yang2021rethinking,yang2021learning}, angle classification~\cite{yang2020arbitrary}, and IoU-aware regression~\cite{yang2019scrdet}.

RSOD backbones commonly improve representation learning through enlarged receptive fields, multi-scale processing, or efficient feature interaction. LSKNet~\cite{Li_2023_ICCV} employs large selective kernels, PKINet~\cite{cai2024poly} uses poly-kernel inception blocks, and LWGANet~\cite{lu2026lwganet} adopts grouped attention. Prior-guided backbones such as LEGNet~\cite{lu2025legnet} and UnravelNet~\cite{LU2026431} further incorporate edge or Gaussian cues. In contrast to applying a uniform prior design across feature levels, SPEANet assigns different fixed operators to different backbone stages and modulates their responses using learned context before fusion.

While predefined filters (e.g., Gaussian, Sobel, Laplacian) provide interpretable, parameter-free inductive biases for structural modeling, their responses are inherently content-agnostic. They are prone to highlighting irrelevant background patterns alongside actual object boundaries. Effective integration therefore necessitates feature-stage-specific operator selection and learned spatial modulation.

% ---------------------------------Approach------------------------------
% ---------------------------------Approach------------------------------
% ---------------------------------Approach------------------------------
\section{Approach}
\subsection{Overview and Unified Formulation}
As illustrated in Figure~\ref{fig_overall}, SPEANet is a four-stage hierarchical backbone with channel dimensions $\{32,64,128,256\}$ and output strides $\{4,8,16,32\}$. The numbers of Hybrid-Prior Refinement Blocks (HPRBs) are $\{1,4,4,2\}$ for Stages I--IV, respectively. Rather than applying one fixed operator uniformly throughout the hierarchy, SPEANet assigns an operator family according to the spatial resolution and representation level of each stage.

For $s\in\{1,2,3,4\}$, $\mathbf{X}^{(s)}\in\mathbb{R}^{C_s\times H_s\times W_s}$ denote the input to Stage $s$. The stem and stage-wise operator families are
\[
\mathcal{K}_{\mathrm{stem}}=\{K_{\mathrm{LoG}}\},\hspace{2mm}
\mathcal{K}^{(s)}=
\begin{cases}
	\{K^S_{\theta}\}_{\theta\in\Theta}\cup\{K^L\}, \hspace{1mm} s=1,\\
	\{K^G_{\sigma},K^L\}, \hspace{3mm} s\in\{2,3,4\}.
\end{cases}
\]
The extractor produces a prior-enhanced response
\[
\mathbf{P}^{(s)}
=\mathcal{E}^{(s)}\!\left(
\mathbf{X}^{(s)};\mathcal{K}^{(s)}
\right).
\]
The prior-enhanced response is not directly propagated. A context-conditioned gate predicts a modulation map $\mathbf{M}^{(s)}$ from the learned feature, and the modulated response is
\[
\widehat{\mathbf{P}}^{(s)}=\mathbf{M}^{(s)}\odot\mathbf{P}^{(s)}.
\]
The single-channel modulation map is broadcast along the channel dimension. The resulting response is aligned with the learned feature through projection, normalization, and residual fusion. This generate-modulate-fuse formulation separates fixed response generation from data-dependent response selection, allowing different structural biases to be introduced without replacing the learned backbone stream.

The grouped convolutions implementing the structural operators use frozen weights and therefore add no trainable operator coefficients. Gradients nevertheless propagate through their responses to preceding learned layers, while the surrounding projections, normalization layers, gates, and interaction branches remain learnable.

\begin{figure}[t]
	\centering
	\includegraphics[width=1.0\columnwidth]{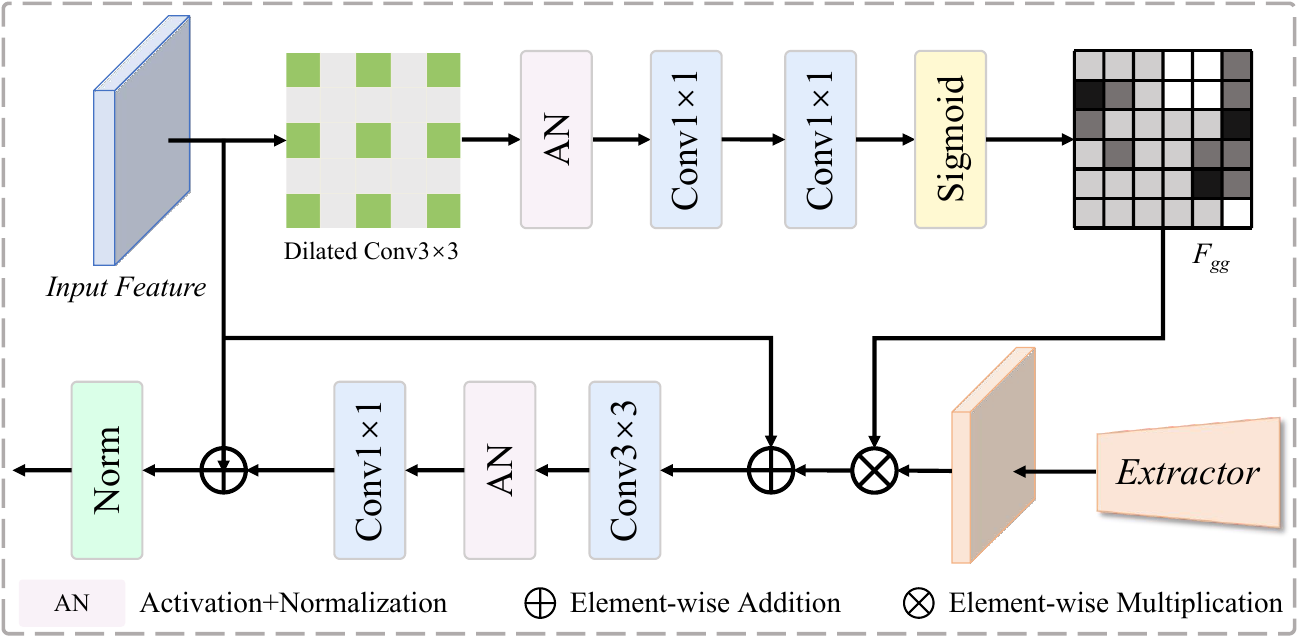}
	\caption{Context-Guided Frequency Filtering (CGFF) module. A dilated context encoder predicts a single-channel spatial gate that modulates the fixed-operator response.}  \label{fig_cgff}
\end{figure}

\subsection{Context-Guided Frequency Filtering (CGFF)}
Fixed operators respond to both object structures and content-irrelevant background patterns. CGFF predicts a spatial gate from the learned feature stream to regulate the corresponding prior-enhanced response. Given a learned feature $\mathbf{X}$ and a stage-specific response $\mathbf{P}$, the context encoder computes
\[
\mathbf{M}=\sigma\!\left(\psi\!\left(\mathcal{C}(\mathbf{X})\right)\right),
\qquad \mathbf{M}\in[0,1]^{1\times H\times W},
\]
where $\mathcal{C}(\cdot)$ consists of a $3\times3$ dilated convolution with dilation 2, followed by normalization, activation, and a $1\times1$ projection, and $\psi(\cdot)$ maps the context feature to a single-channel gate. The modulated response is
\[
\widehat{\mathbf{P}}=\mathbf{M}\odot\mathbf{P}.
\]
CGFF then performs gated residual fusion:
\[
\mathcal{G}(\mathbf{X},\mathbf{P})=
\mathcal{N}\!\left(\mathbf{X}+\mathcal{R}(\mathbf{X}+\widehat{\mathbf{P}})\right),
\]
where $\mathcal{R}(\cdot)$ contains a $3\times3$ convolution, normalization, activation, and a $1\times1$ projection, and $\mathcal{N}(\cdot)$ denotes the output normalization. Because $\mathbf{M}$ is inferred from $\mathbf{X}$ rather than $\mathbf{P}$ alone, the gate uses learned context to control the contribution of potentially content-irrelevant prior responses. CGFF therefore uses learned spatial context to regulate the prior-enhanced branch rather than incorporating it unconditionally.

\subsection{Context-Guided LoG Stem (CLoG-Stem)}
Early downsampling can irreversibly attenuate weak contours before they are encoded by deeper stages. CLoG-Stem therefore introduces a smoothed second-order response before the first resolution reduction. Given an input image $\mathbf{I}$, a $7\times7$ convolution first produces $ \mathbf{X}_0=\phi_7(\mathbf{I}) $.
The LoG branch uses a fixed $7\times7$ kernel with $\sigma=1.0$ and forms a residual contour response
\[
\mathbf{P}_{\mathrm{LoG}}=
\mathcal{N}_2\!\left(
\mathbf{X}_0+\delta\!\left(\mathcal{N}_1(K_{\mathrm{LoG},\sigma}*\mathbf{X}_0)\right)
\right),
\]
where $\delta(\cdot)$ denotes the activation function. The learned feature and LoG-enhanced response are fused by CGFF:
\[
\mathbf{X}_1=\mathcal{G}(\mathbf{X}_0,\mathbf{P}_{\mathrm{LoG}}).
\]
The fused feature is subsequently processed by a depthwise downsampling path and the DRFD module~\cite{lu2023robust} to preserve spatial details during resolution reduction, yielding $\mathbf{X}_{\mathrm{stem}}\in\mathbb{R}^{C_1\times H/4\times W/4}$. The Gaussian in LoG suppresses isolated high-frequency perturbations before second-order differentiation, whereas CGFF attenuates contour responses that are inconsistent with the learned context.

\subsection{Hybrid-Prior Refinement Block (HPRB)}
Each stage consists of repeated HPRBs. Given an input $\mathbf{X}$ at stage $s$, HPRB first selects a stage-specific prior extractor:
\[
\mathcal{E}^{(s)}=
\begin{cases}
	\mathcal{E}_{\mathrm{D}}, & s=1,\\
	\mathcal{E}_{\mathrm{W}}, & s\in\{2,3,4\},
\end{cases}
\]
where $\mathcal{E}_{\mathrm{D}}$ and $\mathcal{E}_{\mathrm{W}}$ denote MDP-FASE and WASM, respectively. The block then computes
\[
\mathbf{P}=\mathcal{E}^{(s)}(\mathbf{X}),\qquad
\mathbf{Z}=\mathcal{G}(\mathbf{X},\mathbf{P}),\qquad
\mathbf{U}=\mathcal{S}(\mathbf{Z}),
\]
where $\mathcal{G}$ and $\mathcal{S}$ are CGFF and FASE modules. Stochastic depth is applied through
\[
\mathbf{Y}=\mathbf{X}+\operatorname{DropPath}(\mathbf{U}-\mathbf{X}).
\]
Here, $\mathbf{U}-\mathbf{X}$ represents the complete residual transformation of HPRB; dropping this branch reduces the block to the identity mapping. This formulation preserves an identity path across the block while applying prior extraction, context-conditioned modulation, and frequency-aware encoding within the residual branch. MDP-FASE is assigned to Stage I because its directional bank operates on spatially detailed features, whereas WASM is applied to Stages II--IV, as its approximation-detail interaction better suits broader contexts and larger channel dimensions of deeper layers.

\begin{figure}[t]
	\centering
	\includegraphics[width=1.0\columnwidth]{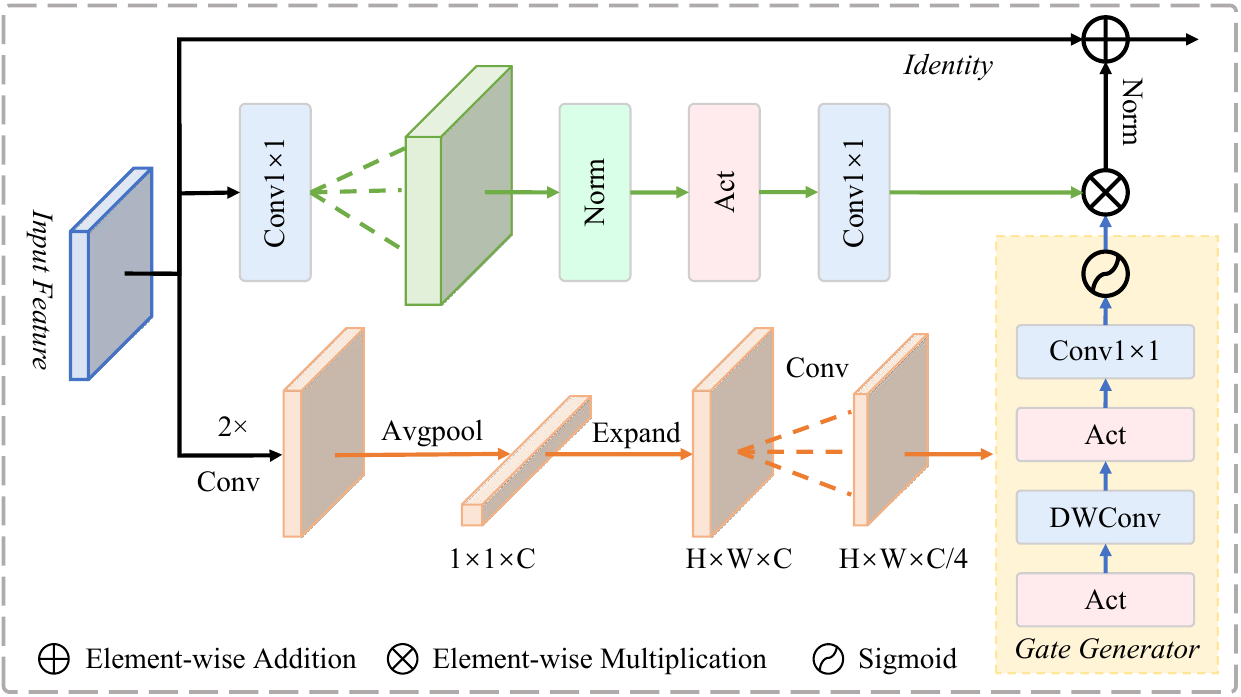}
	\caption{Frequency-Aware Scaling and Encoding (FASE) module. A globally aggregated descriptor conditions the scaling of the pointwise-transformed feature branch.}	\label{fig_fase}
\end{figure}

\subsubsection{Frequency-Aware Scaling and Encoding (FASE)}
FASE replaces a conventional pointwise feed-forward transformation with globally conditioned response scaling. Given $\mathbf{X}\in\mathbb{R}^{C\times H\times W}$, it applies a depthwise $3\times3$ convolution and a $1\times1$ projection, followed by global average pooling:
\[
\mathbf{q}=\operatorname{GAP}\!\left(\phi_g(\operatorname{DWConv}_{3\times3}(\mathbf{X}))\right),
\qquad \mathbf{q}\in\mathbb{R}^{C\times1\times1}.
\]
The global descriptor is broadcast to the original resolution and transformed into a single-channel scaling map:
\[
\mathbf{A}=\sigma\!\left(\psi_f(\operatorname{Broadcast}(\mathbf{q}))\right),
\qquad \mathbf{A}\in[0,1]^{1\times H\times W}.
\]
Here, $\psi_f$ consists of a pointwise projection, activation, depthwise $3\times3$ filtering, a second activation, and a final pointwise projection. Since $\mathbf{A}$ is generated solely from the globally pooled descriptor $\mathbf{q}$, it provides global response conditioning rather than location-specific attention. In parallel, a two-layer pointwise projection $\Phi(\cdot)$ transforms the input features. The FASE output is
\[
\mathcal{S}(\mathbf{X})=\mathbf{X}+\mathcal{N}\!\left(\Phi(\mathbf{X})\odot\mathbf{A}\right).
\]
The global descriptor summarizes the response distribution of the current feature map, while $\mathbf{A}$ controls the magnitude of the transformed branch before residual addition. The term frequency-aware refers to conditioning on responses summarized after depthwise spatial filtering; FASE does not explicitly estimate entropy or perform a spectral transform.

\subsubsection{Multi-Order Directional Prior with FASE (MDP-FASE)}
Stage I retains sufficient spatial resolution for derivative responses to describe localized object structures. MDP-FASE therefore combines a four-direction Sobel bank with an isotropic Laplacian response. Let $\Theta=\{0^{\circ},45^{\circ},90^{\circ},135^{\circ}\}$. The first-order response bank is
\[
\mathbf{S}=\operatorname{Concat}_{\theta\in\Theta}
\left(K^S_{\theta}*\mathbf{X}\right)
\in\mathbb{R}^{4C\times H\times W}.
\]
A two-layer pointwise transform $\phi_d$ compresses the directional bank to $C$ channels. Channel-wise recalibration is then obtained from the globally pooled response:
\[
\mathbf{U}=\phi_d(\mathbf{S}),\qquad
\mathbf{w}=\sigma\!\left(\phi_w(\operatorname{GAP}(\mathbf{U}))\right),\qquad
\mathbf{D}_1=\mathbf{U}\odot\mathbf{w}.
\]
The second-order branch applies an eight-neighborhood Laplacian kernel:
\[
\mathbf{D}_2=\mathcal{N}_L(K^L*\mathbf{X}).
\]
The two derivative orders are concatenated, encoded by FASE, and projected back to $C$ channels:
\[
\mathcal{E}_{\mathrm{D}}(\mathbf{X})=
\Pi_{\mathrm{D}}\!\left(
\mathcal{S}\!\left([\mathbf{D}_1,\mathbf{D}_2]\right)
\right).
\]
where $\mathcal{N}_L$ denotes normalization and $\Pi_{\mathrm{D}}$ is a pointwise projection. The directional bank captures orientation-sensitive first-order variations, whereas the Laplacian captures isotropic second-order changes. Their combination provides complementary local responses without the need for learning the derivative kernels themselves. The learned transforms adapt the fixed responses to the feature channels required by the detector.

\subsubsection{Wavelet-Inspired Analysis-Synthesis Mixer (WASM)}
WASM follows the approximation-detail interaction principle of wavelet analysis without performing an explicit discrete wavelet transform. Given an input $\mathbf{X}$, WASM constructs two learnable analysis branches guided by fixed low-pass and high-pass operators:
\[
\begin{aligned}
	\mathbf{A}
	&=\delta\!\left(
	\mathcal{N}_a\!\left(\phi_{\mathrm{A}}(\mathbf{X})+K^G_{\sigma}*\mathbf{X}\right)
	\right),\\
	\mathbf{D}
	&=\delta\!\left(
	\mathcal{N}_d\!\left(\phi_{\mathrm{D}}(\mathbf{X})+K^L*\mathbf{X}\right)
	\right),
\end{aligned}
\]
where $K^G_{\sigma}$ is a fixed $5\times5$ Gaussian kernel with $\sigma=1.0$, $K^L$ is the Laplacian kernel, and $\phi_{\mathrm{A}}$ and $\phi_{\mathrm{D}}$ are depthwise-separable transforms. The approximation and detail branches modulate each other through
\[
\widetilde{\mathbf{A}}=\mathbf{A}\odot\sigma(W_d(\mathbf{D})),
\qquad
\widetilde{\mathbf{D}}=\mathbf{D}\odot\sigma(W_a(\mathbf{A})),
\]
where $W_a$ and $W_d$ are $1\times1$ projections. The synthesis operation concatenates the modulated branches and restores the original channel dimension:
\[
\mathcal{E}_{\mathrm{W}}(\mathbf{X})=
\mathbf{X}+\mathcal{N}_{\mathrm{W}}\!\left(
W_f\!\left([\widetilde{\mathbf{A}},\widetilde{\mathbf{D}}]\right)
\right).
\]
The approximation branch supplies broader low-frequency context to the detail branch, while the detail branch conditions the preservation of local variations in the approximation branch. This bidirectional interaction prevents the high-frequency response from being treated independently of its surrounding context and is therefore applied to the deeper feature stages.

% ---------------------------------Experiments------------------------------
% ---------------------------------Experiments------------------------------
% ---------------------------------Experiments------------------------------
\section{Experiments}
We evaluate SPEANet from four perspectives. First, comparisons with representative RSOD methods assess the accuracy-parameter trade-off. Second, cross-detector experiments examine whether the backbone improvements transfer across one-stage and two-stage detection frameworks. Third, evaluations on DOTA-v1.5, DIOR-R, SODA-A, and MODA test generalization across object scales, scene distributions, and input modalities. Finally, ablation and qualitative analyses examine the stage-specific configuration and the spatial modulation produced by CGFF.

\subsection{Datasets and Experimental Setup}
We evaluate SPEANet on five benchmarks: DOTA-v1.0~\cite{xia2018dota}, DOTA-v1.5~\cite{xia2018dota}, DIOR-R~\cite{cheng2022anchor}, SODA-A~\cite{cheng2023towards}, and MODA~\cite{han2026moda}. These datasets cover complementary RSOD settings, including arbitrary orientations, dense clutter, extreme scale variations, tiny instances, and multispectral inputs. All backbones are initialized with ImageNet-1K pretraining, and the detectors are fine-tuned using standard training schedules (3$\times$). More dataset statistics and implementation details are provided in the Supplementary Material.

\subsection{Quantitative Results}

\begin{table}[ht]
	\centering
	\scriptsize
	\setlength{\tabcolsep}{4pt}
	\resizebox{\columnwidth}{!}{
		\begin{tabular}[c]{l|c|c|c}
			\toprule
			\multicolumn{1}{c|}{\textbf{Method}} 
			& \textbf{Backbone} 
			& \textbf{\#P$\downarrow$} 
			& \textbf{mAP$\uparrow$}\\
			\midrule
			\multicolumn{4}{l}{\textit{One Stage}}\\
			\midrule
			R$^3$Det~\cite{yang2021r3det} & ResNet-50~\cite{he2016deep} & 41.9M & 69.70\\
			SASM~\cite{hou2022shape} & ResNet-50~\cite{he2016deep} & 36.6M & 74.92 \\
			O-RepPoints~\cite{li2022oriented} & ResNet-50~\cite{he2016deep} & 36.6M & 75.97 \\
			R$^3$Det-GWD~\cite{yang2021rethinking} & ResNet-50~\cite{he2016deep} & 41.9M & 76.34 \\
			R$^3$Det-KLD~\cite{yang2021learning} & ResNet-50~\cite{he2016deep} & 41.9M & 77.36 \\
			\midrule
			\multirow[c]{4}{*}{S$^2$ANet~\cite{han2021align}} & ResNet-50~\cite{he2016deep} & 38.5M & 74.12 \\
			& ARC-R50~\cite{pu2023adaptive} & 71.8M & 75.49 \\
			& PKINet-S~\cite{cai2024poly} & 24.8M & 77.83 \\
			& \cellcolor{gray!25}\textbf{SPEANet} & \cellcolor{gray!25}\textbf{14.0M} & \cellcolor{gray!25}78.16 \\
			\midrule
			\multicolumn{4}{l}{\textit{Two Stage}}\\
			\midrule
			CenterMap~\cite{CenterMap} & ResNet-50~\cite{he2016deep} & 41.1M & 71.59 \\
			SCRDet~\cite{yang2019scrdet} & ResNet-50~\cite{he2016deep} & 41.9M & 72.61 \\
			FR-O~\cite{ren2016faster} & ResNet-50~\cite{he2016deep} & 41.1M & 73.17 \\
			Roi Trans.~\cite{ding2019learning} & ResNet-50~\cite{he2016deep} & 55.1M & 74.05 \\
			G.V.~\cite{xu2020gliding} & ResNet-50~\cite{he2016deep} & 41.1M & 75.02 \\
			ReDet~\cite{han2021redet} & ResNet-50~\cite{he2016deep} & 31.6M & 76.25 \\
			\midrule
			\multirow[c]{5}{*}{O-RCNN~\cite{xie2021oriented}} 
			& ResNet-50~\cite{he2016deep} & 41.1M & 75.87 \\
			& ARC-R50~\cite{pu2023adaptive} & 74.4M & 77.35 \\
			& LSKNet-S~\cite{Li_2023_ICCV} & 31.0M & 77.49 \\
			& PKINet-S~\cite{cai2024poly} & 30.8M & \underline{78.39} \\
			& \cellcolor{gray!25}\textbf{SPEANet} & \cellcolor{gray!25}\underline{23.0M} & \cellcolor{gray!25}\textbf{78.55} \\
			\bottomrule
		\end{tabular}
	}
	\caption{Performance comparison on the DOTA-v1.0 test set under single-scale training and testing. The best results are shown in bold, and the second-best results are underlined. }% Results for other methods are sourced from LEGNet~\cite{lu2025legnet}.
	\label{tab:result_dota1.0}
\end{table}

\begin{table}[ht]
	\centering
	\scriptsize
	\setlength{\HeaderShift}{\dimexpr-\aboverulesep/2-\belowrulesep/2-\cmidrulewidth/2\relax}
	\resizebox{\linewidth}{!}{
		\begin{tabular}{c|l|cc|>{\columncolor{gray!25}}c}
			\toprule
			\multicolumn{2}{c|}{\multirow{2}{*}{Framework}} & ResNet-50 & PKINet-S & \textbf{SPEANet} \\
			\multicolumn{2}{c|}{} & (23.28M) & (13.70M) & \textbf{(5.97M)} \\
			\midrule
			\multirow[c]{3}{*}{\textit{One Stage}} 
			& R-FCOS~\cite{tian2020fcos} & 72.45 & 74.86 & \textbf{76.27} \\
			& R$^3$Det~\cite{yang2021r3det} & 69.70 & 75.89 & \textbf{76.32}\\
			& S$^2$ANet~\cite{han2021align} & 74.12 & 77.83 & \textbf{78.16}\\
			\midrule
			\multirow[c]{4}{*}{\textit{Two Stage}}
			& FR-O~\cite{ren2016faster} & 73.17 & 76.45 & \textbf{76.86}\\
			& Roi Trans.~\cite{ding2019learning} & 74.05 & 77.17 & \textbf{79.15} \\
			& O-RCNN~\cite{xie2021oriented} & 75.87 & 78.39 & \textbf{78.55}\\
			& CTRP~\cite{Sun2025CTRP} & 77.17 & 78.38 & \textbf{79.85}\\ 
			\bottomrule
		\end{tabular}
	}
	\caption{Comparison of mAP across different detectors and backbones (ResNet-50 and PKINet-S) on the DOTA-v1.0 test set. Only backbone parameters are included. }% Results for other methods are sourced from LEGNet~\cite{lu2025legnet}.
	\label{tab:result_dota10_diffdetector}
\end{table}

\subsubsection{Performance on DOTA-v1.0}
As shown in Table~\ref{tab:result_dota1.0}, SPEANet achieves a competitive accuracy-parameter trade-off. SPEANet with S$^2$ANet~\cite{han2021align} reaches 78.16\% mAP with 14M total parameters, and SPEANet with O-RCNN~\cite{xie2021oriented} reaches 78.55\% mAP with 23.0M total parameters. These results outperform PKINet-S~\cite{cai2024poly} under both detector settings while using fewer parameters. The gains are observed on a benchmark containing small, elongated, and densely distributed objects, for which preserving localized structural cues is important.

Table~\ref{tab:result_dota10_diffdetector} compares SPEANet with ResNet-50~\cite{he2016deep} and PKINet-S~\cite{cai2024poly} under seven detection frameworks on DOTA-v1.0. 
The improvement margins vary with the detector architecture, but SPEANet outperforms both comparison backbones in all seven settings. Relative to PKINet-S, the gains are 1.41, 0.43, and 0.33 points for R-FCOS~\cite{tian2020fcos}, R$^3$Det~\cite{yang2021r3det}, and S$^2$ANet, respectively, and 0.41, 1.98, 0.16, and 1.47 points for FR-O~\cite{ren2016faster}, Roi Trans.~\cite{ding2019learning}, O-RCNN, and CTRP~\cite{Sun2025CTRP}. The consistent improvements across anchor-free, refinement-based, and two-stage frameworks indicate that the learned representation is not specialized to one detection head.

\begin{table}[t]
	\centering	
	\scriptsize
	\setlength{\tabcolsep}{3pt}
	\resizebox{\columnwidth}{!}{
		\begin{tabular}[c]{l|c|cccc|c}
			\toprule
			\multicolumn{1}{c|}{\textbf{Method}} & \textbf{\#P$\downarrow$} 
			& LV & SH & BC & HC & \textbf{mAP$\uparrow$}\\
			\midrule
			RetinaNet-O~\cite{lin2017focal} & 36.4M & 56.79 & 73.31 & 76.02 & 48.06 & 59.16 \\
			FR-O~\cite{ren2016faster} & 41.1M & 68.98 & 79.37 & 77.38 & 60.47 & 62.00 \\
			Mask R-CNN~\cite{he2017piotr} & 44.4M & 71.34 & 79.75 & 74.21 & 57.81 & 62.67 \\
			HTC~\cite{chen2019hybrid} & 77.5M & 73.31 & 80.31 & 75.12 & 55.87 & 63.40 \\
			ReDet~\cite{han2021redet} & 31.6M & 75.73 & 80.92 & 75.81 & 63.33 & 66.86 \\
			LSKNet-S~\cite{Li_2023_ICCV} & 31.0M & 77.45 & 81.17 & 79.44 & 55.81 & 70.26 \\
			SOOD~\cite{xi2024structure} & - & 76.90 & 86.97 & 78.62 & 55.97 & 70.39 \\
			SPCNet~\cite{zheng2025spcnet} & 25.3M & 76.66 & 87.35 & 78.65 & 61.25 & 70.60 \\
			PKINet-S~\cite{cai2024poly} & 30.8M & 76.85 & 88.38 & 79.04 & 64.07 & 71.47 \\
			\rowcolor{gray!25}
			\textbf{SPEANet} & \textbf{23.0M} & \textbf{82.00} & \textbf{88.61} & \textbf{83.42} & \textbf{74.75} & \textbf{72.24} \\
			\bottomrule
		\end{tabular}
	}
	\caption{Performance comparison on the DOTA-v1.5 test set. Representative categories involving small objects, elongated structures, or cluttered backgrounds are reported; complete category results are provided in Supplementary Material.}	\label{tab:result_dota15}
\end{table}

\begin{table}[t]
	\centering
	\scriptsize
	\setlength{\tabcolsep}{2pt}
	\setlength{\HeaderShift}{\dimexpr-\aboverulesep/2-\belowrulesep/2-\cmidrulewidth/2\relax}
	\resizebox{\linewidth}{!}{
		\begin{tabular}{l|c|cc|c}
			\toprule
			\multicolumn{1}{c|}{\multirow{2}{*}[\HeaderShift]{\textbf{Method}}}
			& \multirow{2}{*}[\HeaderShift]{Backbone}
			& \multicolumn{2}{c|}{Overall / Backbone}
			& \multirow{2}{*}[\HeaderShift]{\textbf{mAP (\%)$\uparrow$}} \\
			\cmidrule(lr){3-4}
			& & \textbf{\#P$\downarrow$}(M)  & FLOPs(G) &  \\
			\midrule
			FR-O~\cite{ren2016faster} & ResNet50 & 41.1 / 23.3 & 148.2 / 52.4& 59.54 \\
			G.V.~\cite{xu2020gliding}& ResNet50 & 41.1 / 23.3  & 148.2 / 52.4 & 60.06 \\
			Roi Trans.~\cite{ding2019learning} & ResNet50 & 55.1 / 23.3  & 176.2 / 52.4 & 63.87 \\
			LSKNet~\cite{Li_2023_ICCV}  & LSKNet-S & 31.0 / 13.8 & 125.1 / 32.9 & 65.90 \\
			Oriented Rep~\cite{li2022oriented} & ResNet50 & 36.6 / 23.3 &  118.7 / 52.4 & 66.71 \\
			DCFL~\cite{xu2023dynamic} & ResNet50 & 36.1 / 23.3 & - / - & 66.80 \\
			PKINet~\cite{cai2024poly} & PKINet-S & 30.8 / 13.7 & 131.8 / 39.7 & 67.03 \\
			\rowcolor{gray!25}
			\textbf{SPEANet} & SPEANet &\textbf{23.0} / \textbf{6.0} & \textbf{107.5} / \textbf{16.0} & \textbf{67.30} \\
			\bottomrule
		\end{tabular}
	}
	\caption{Experimental results on the DIOR-R test set. \#P  and FLOPs were measured at a $800 \times 800$ resolution.}
	\label{tab:result_diorr}
\end{table}

\subsubsection{Performance on DOTA-v1.5}
As shown in Table~\ref{tab:result_dota15}, SPEANet with O-RCNN achieves 72.24\% mAP, exceeding LSKNet-S and PKINet-S by 1.98 and 0.77 percentage points, respectively. It also ranks first or second in 6 of 16 categories and improves helicopter (HC) AP by 10.68 points over the strongest compared baseline. These results indicate that SPEANet remains effective on DOTA-v1.5, which includes more challenging tiny instances and an additional container crane category.

\subsubsection{Performance on DIOR-R}
As shown in Table~\ref{tab:result_diorr}, SPEANet with O-RCNN achieves 67.30\% mAP with 23.0M total parameters, outperforming PKINet-S by 0.27 percentage points while using 7.7M fewer backbone parameters. It also requires only 16.0G backbone FLOPs under the reported setting. The improvement on DIOR-R, whose categories and scene distributions differ from DOTA, indicates generalization of the proposed backbone beyond a single benchmark.

\begin{table}[!t]
	\centering	
	\scriptsize
	\setlength{\tabcolsep}{2pt}
	\resizebox{\linewidth}{!}{
		\begin{tabular}[c]{l|ccc|cc}
			\toprule
			\multicolumn{1}{c|}{\textbf{Method}} & AP$_{50:95}$ & AP$_{50}$ & AP$_{75}$ & \textbf{\#P$\downarrow$}  & FLOPs\\ 
			\midrule
			Rotated Faster RCNN~\cite{ren2016faster} &  32.5 & 70.1 & 24.3 & 41.1M & 292G\\
			Rotated RetinaNet~\cite{lin2017focal} &  26.8 & 63.4 & 16.2 & 36.2M & 800G\\
			RoI Trans.~\cite{ding2019learning} &  36.0 & 73.0 & 30.1 & 55.1M & 306G\\
			Gliding Vertex~\cite{xu2020gliding} &  31.7 & 70.8 & 22.6 & 41.1M & 292G\\
			Oriented RCNN~\cite{xie2021oriented} &  34.4 & 70.7 & 28.6 & 41.1M & 292G\\
			S$^2$A-Net~\cite{han2021align} & 28.3 & 69.6 & 13.1 & 38.6M & 733G\\
			DODet~\cite{cheng2022dual} & 31.6 & 68.1 & 23.4 & 69.3M & 555G\\
			Oriented RepPoints~\cite{li2022oriented} & 26.3 & 58.8 & 19.0 & 55.7M & 827G\\
			DHRec~\cite{nie2022multi} & 30.1 & 68.8 & 19.8 & 32.0M & 793G\\
			\rowcolor{gray!25}
			\textbf{SPEANet} & \textbf{37.4} & \textbf{74.5} & \textbf{32.9}& \textbf{23.0M} & \textbf{198G}\\
			\bottomrule
		\end{tabular}
	}
	\caption{Comparison on the SODA-A test set. \#P and FLOPs are measured at $1200 \times 1200$ resolution. }%AP$_{50:95}$ denotes COCO-style mean AP over IoU thresholds from 0.50 to 0.95. Results of compared methods follow SODA-Benchmark~\cite{cheng2023towards}.
	\label{tab:result_sodaa}
\end{table}

\subsubsection{Performance on SODA-A}
SODA-A is a particularly challenging aerial detection benchmark dominated by tiny objects, whose limited pixels, weak boundaries, and dense distributions make both recognition and localization difficult. As shown in Table~\ref{tab:result_sodaa}, SPEANet with O-RCNN achieves 37.4\% AP$_{50:95}$, 74.5\% AP$_{50}$, and 32.9\% AP$_{75}$ with 23.0M total parameters. Compared with the O-RCNN baseline, SPEANet improves these metrics by 3.0, 3.8, and 4.3 percentage points, respectively, while using 18.1M fewer parameters. In particular, the larger improvement at AP$_{75}$ indicates that the gain persists under a stricter localization criterion. These results suggest that preserving structural cues in high-resolution features benefits the detection and localization of tiny objects without increasing model capacity.

\begin{table}[!t]
	\centering	
	\scriptsize
	\setlength{\tabcolsep}{2pt}
	\resizebox{\linewidth}{!}{
		\begin{tabular}[c]{l|ccc|cc}
			\toprule
			\multicolumn{1}{c|}{\textbf{Method}} & {AP$_{50}$} & {AP$_{75}$} & mAP  & \textbf{\#P$\downarrow$}   & FLOPs\\
			\midrule
			R$^3$Det~\cite{yang2021r3det}  & 60.2 & 36.7 & 34.8 & 42.0M & 362.2G\\
			S$^2$ANet~\cite{han2021align} & 63.1 & 38.7 & 36.7 & 38.8M & 216.5G\\
			R-FCOS~\cite{tian2020fcos} & 63.5 & 41.8 & 39.1 & 32.2M & 226.9G\\
			RoI Trans.~\cite{ding2019learning} & 65.4 & 43.4 & 40.7 & 55.3M & 244.7G\\
			StripRCNN~\cite{yuan2026strip} & 66.1 & 44.0 & 41.0 & 45.2M & 231.8G\\
			CFA~\cite{guo2021beyond} & 66.2 & 43.2 & 40.6 & 36.9M & 213.6G\\
			OSSDet~\cite{han2026moda} & 69.0 & 45.9 & 42.7 & 36.5M & 263.1G\\
			\rowcolor{gray!25}
			\textbf{SPEANet} & \textbf{69.8} & \textbf{47.1} & \textbf{43.8} & \textbf{23.0M} & \textbf{156.5G}\\
			\bottomrule
		\end{tabular}
	}
	\caption{Comparison with representative methods on the MODA test set. FLOPs are measured at an input resolution of $1200\times900$; \#P denotes the total number of parameters.}
	\label{tab:result_moda}
\end{table}

\subsubsection{Performance on MODA}
As shown in Table~\ref{tab:result_moda}, SPEANet with O-RCNN achieves 69.8\% AP$_{50}$ and 43.8\% mAP, exceeding OSSDet~\cite{han2026moda} by 0.8 and 1.1 points, with 23.0M total parameters. The result demonstrates that the proposed backbone can also be applied to eight-channel multispectral inputs after adapting the input projection.

Taken together, the five benchmarks probe complementary aspects of the proposed design. DOTA-v1.0 and DOTA-v1.5 contain dense and arbitrarily oriented instances, SODA-A emphasizes tiny objects, DIOR-R broadens the category and scene distribution, and MODA replaces RGB input with eight-channel multispectral data. SPEANet uses fewer parameters than the compared backbones while improving the reported detection metrics in each setting. These results do not isolate the effect of every individual operator, but they show that the stage-specific backbone design remains effective beyond a single detector, dataset, or input modality.

\subsection{Qualitative Results}
\begin{figure}[t]	\centering
	\includegraphics[width=1\columnwidth]{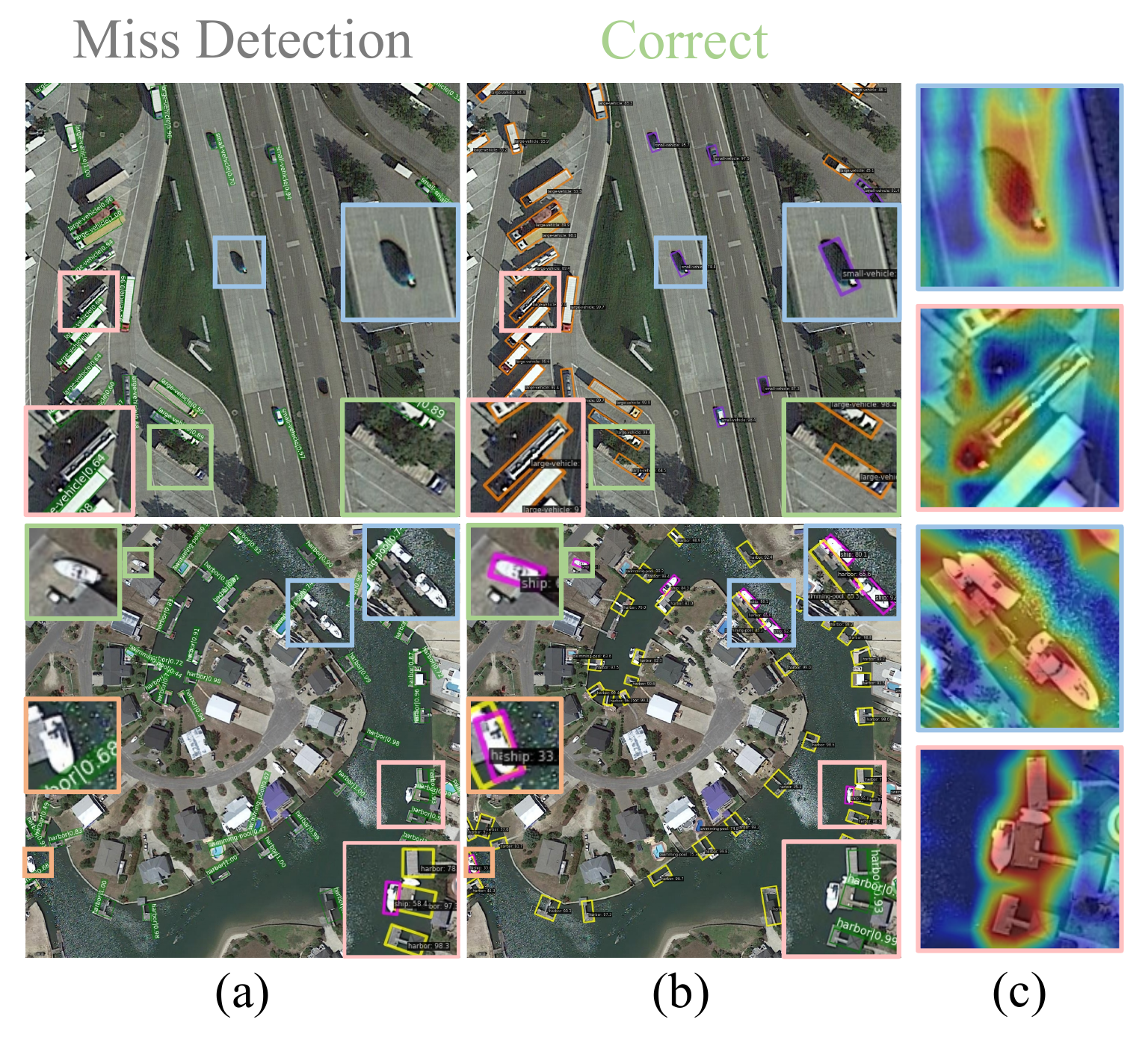}
	\caption{Detection results and CGFF gating maps on the DOTA-v1.0 test set. \textbf{(a)} PKINet-S. \textbf{(b)} SPEANet. \textbf{(c)} Spatial gating maps produced by CGFF.}
	\label{fig_vis2}  
\end{figure}

Figure~\ref{fig_vis2} compares SPEANet and PKINet-S on representative DOTA-v1.0 scenes. SPEANet recovers several ships and small vehicles that are missed by the comparison backbone and reduces false positives around repetitive background structures. CGFF maps visualize the single-channel gate $\mathbf{M}$. As visualized in Figure~\ref{fig_vis2}(c), CGFF assigns high gate values (red) to structural boundaries of ships and vehicles, while actively suppressing responses over homogeneous water surfaces and regular background textures (blue). The maps assign relatively high gate values to object boundaries and lower values to homogeneous or repetitive backgrounds. This behavior is consistent with CGFF using learned context to modulate content-agnostic prior responses.

\begin{table}[t]
\centering
\scriptsize
\setlength{\tabcolsep}{2pt}
\setlength{\HeaderShift}{\dimexpr-\aboverulesep/2-\belowrulesep/2-\cmidrulewidth/2\relax}
\resizebox{\linewidth}{!}{
\begin{tabular}{ccc|c|c|ccc}
\toprule
\multirow{2}{*}[\HeaderShift]{CLoG-Stem} &
\multirow{2}{*}[\HeaderShift]{MDP-FASE} &
\multirow{2}{*}[\HeaderShift]{WASM} &
\multirow{2}{*}[\HeaderShift]{\textbf{\#P$\downarrow$}} &
\multirow{2}{*}[\HeaderShift]{\textbf{mAP$\uparrow$}} &
\multicolumn{3}{c}{Category AP} \\
\cmidrule(lr){6-8}
& & & & & SV & LV & SP \\
\midrule
\ding{51} &           &           & 5.351M & 76.581 & 78.75 & 84.51 & 72.81 \\
& \ding{51} &           & 6.526M & 77.184 & 78.43 & 85.60 & 73.23 \\
&           & \ding{51} & \textbf{4.786M} & 77.605 & 78.53 & 84.53 & 78.62 \\
\ding{51} & \ding{51} &           & 6.531M & 78.054 & 79.38 & 85.15 & 80.24 \\
\ding{51} &           & \ding{51} & 4.791M & 77.778 & 78.86 & 84.92 & 78.99 \\
& \ding{51} & \ding{51} & 5.966M & 78.022 & 79.96 & 85.51 & 73.52 \\
\rowcolor{gray!25}
\multicolumn{3}{c|}{\textbf{SPEANet}} &
5.971M & \textbf{78.550} &
\textbf{80.92} & \textbf{85.86} & \textbf{80.96} \\
\bottomrule
\end{tabular}
}
\caption{Ablation study on DOTA-v1.0. Variants containing only MDP-FASE or WASM apply the corresponding module to all backbone stages. When both are enabled, MDP-FASE is used in Stage I and WASM in Stages II--IV. All variants use 100-epoch ImageNet-1K pretraining and Oriented R-CNN. \#P denotes backbone parameters. SV, LV, and SP denote Small Vehicle, Large Vehicle, and Swimming Pool. Complete results are provided in Supplementary Material.}
\label{tab:ablation}
\end{table}

\subsection{Ablation Studies}
Table~\ref{tab:ablation} evaluates the three principal components of SPEANet. When CLoG-Stem is disabled, it is replaced by a learnable $4\times4$ stride-4 convolution. Disabled MDP-FASE or WASM modules are replaced by convolutional blocks with the same input and output dimensions. Variants containing only MDP-FASE or WASM apply that module throughout the backbone, whereas the complete model assigns MDP-FASE to Stage I and WASM to Stages II--IV. All variants use 100-epoch ImageNet-1K pretraining and Oriented R-CNN on DOTA-v1.0.

SPEANet achieves 78.550\% mAP with 5.971M backbone parameters. Adding CLoG-Stem to the stage-specific MDP-FASE+WASM variant improves mAP from 78.022\% to 78.550\%. Removing MDP-FASE or WASM from the complete configuration reduces mAP by 0.772 and 0.496 percentage points, respectively. These results demonstrate the complementary contributions of the three components. The complete model also achieves the highest AP on the reported SV, LV, and SP categories.

Compared with applying either MDP-FASE or WASM throughout the backbone, their stage-specific combination achieves higher accuracy while retaining a compact parameter budget. This result supports using directional priors at the high-resolution first stage and the more parameter-efficient approximation-detail interaction in deeper stages.

\subsection{Discussion}
The results indicate that fixed structural operators complement learned RSOD features when their responses are adapted to feature stage and image content. The ablation results further suggest that assigning MDP-FASE to Stage I and the more compact WASM to deeper stages provides a favorable accuracy-parameter trade-off, while cross-detector and multi-dataset evaluations support compatibility with different detection heads and input modalities. Nevertheless, parameter count and FLOPs do not directly reflect hardware latency. Latency-aware evaluation remains future work.

% ---------------------------------Conclusion------------------------------
% ---------------------------------Conclusion------------------------------
% ---------------------------------Conclusion------------------------------
\section{Conclusion}
We presented SPEANet, a parameter-efficient RSOD backbone that integrates fixed structural operators through stage-specific modeling and context-conditioned modulation. CLoG-Stem preserves smoothed contour responses before early downsampling, MDP-FASE models multi-order directional variations at high resolution, and WASM performs approximation-detail interaction in deeper stages. Experiments on five benchmarks and seven detection frameworks demonstrate a favorable accuracy-parameter trade-off. These results establish stage-specific operator modeling as an effective complement to learned hierarchical representations.

\setcounter{page}{1}
\appendix

% 使用 \twocolumn 的可选参数 []，中括号内的内容会横跨双栏居中显示在页面顶部
\twocolumn[
\begin{center}
	\Large \bfseries % 设置字体大小和加粗
	Supplementary Material for ``SPEANet: Structural Prior Enhanced Attention Network for Parameter-Efficient Remote Sensing Object Detection''
	\vspace{3ex} % 标题下方留出一些空白
\end{center}
]

This supplement provides the additional materials supporting the main paper ``SPEANet: Structural Prior Enhanced Attention Network for Parameter-Efficient Remote Sensing Object Detection''. We present a detailed overview of the SPEANet architecture; descriptions of the five public benchmark datasets; the experimental protocols used for object detection; extended quantitative results and ablation studies; additional qualitative visualizations on four benchmark datasets; and the design and implementation details of the key components. Together, these materials provide further insights into the proposed method and facilitate the reproducibility of our experiments.

\section{SPEANet Architecture Overview}
\begin{figure}[t] \centering
\includegraphics[width=1.0\columnwidth]{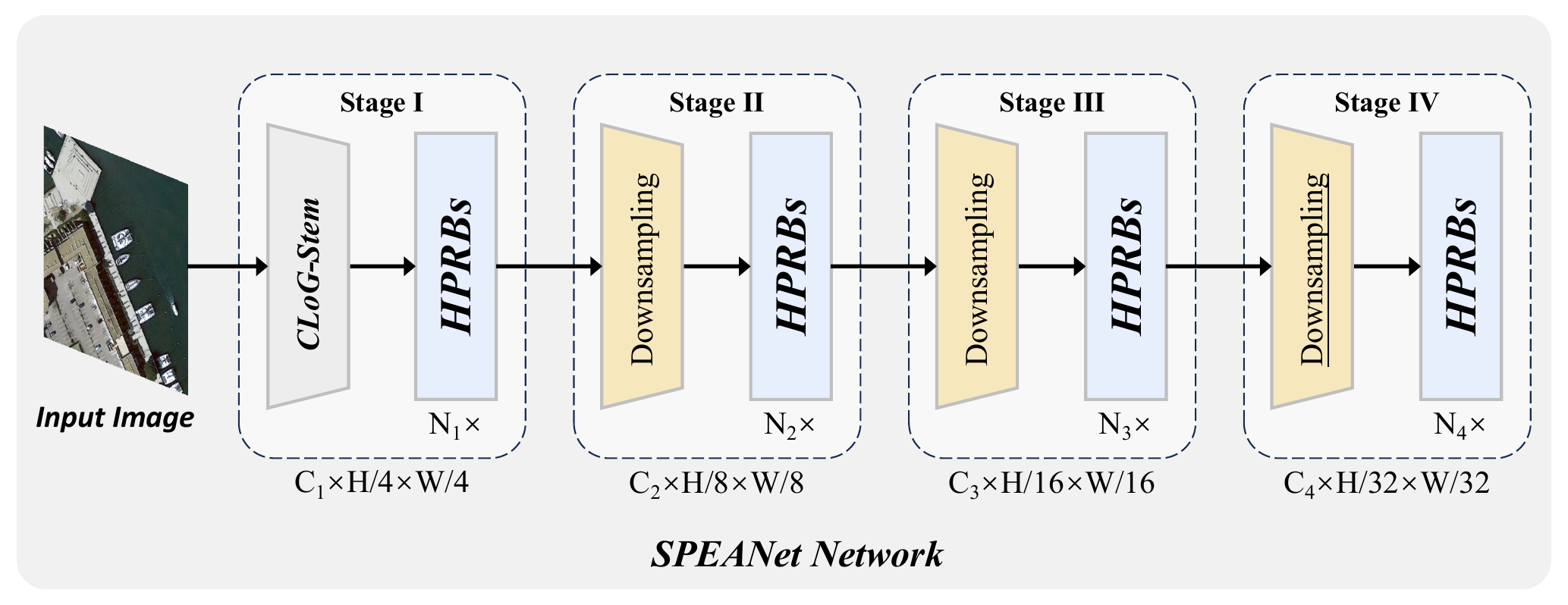}
\caption{Detailed overview of the SPEANet architecture} \label{fig_overall_vis}  
\end{figure}

As shown in Figure~\ref{fig_overall_vis}, SPEANet adopts a four-stage hierarchical architecture composed of repeated Hybrid-Prior Refinement Blocks (HPRBs). Table~\ref{tab:speanet_configuration} presents its configuration. 

\subsubsection{Channel Dimensions:} In the SPEANet backbone, the input channel number is initially set to 3 (for the MODA dataset~\cite{han2026moda}, it is adjusted to 8 to receive multispectral inputs). The output channel dimensions of the four stages are set to 32, 64, 128, and 256, respectively.

\subsubsection{Downsampling Ratio:} The stem layer downsamples the input to 1/4 resolution, and the transitions to Stages II--IV further downsample the feature maps by a factor of 2. The four stages therefore produce feature maps at 1/4, 1/8, 1/16, and 1/32 resolutions, respectively.

\subsubsection{Configuration:} In the SPEANet backbone, different building modules are adopted across different stages. The MDP-FASE module is employed in Stage I, whereas the WASM module is used in Stages~II-IV. Here, $N_i$ represents the number of blocks stacked in the i-th stage.

The parameter allocation across different stages follows the characteristics of hierarchical feature learning. In the early stage, feature maps preserve high spatial resolution with relatively fewer channels; therefore, MDP-FASE is employed to capture detailed structural information. As the network goes deeper, the channel dimensions increase and lead to higher parameter costs. To maintain overall parameter efficiency, WASM adopts lightweight 1$\times$1 convolutions for channel projection and feature interaction, reducing additional parameters while effectively preserving approximation-detail information exchange.

\begin{table}[t] \centering \small
\resizebox{\linewidth}{!}{
\begin{tabular}{c|c|c|c} \toprule
$\{C_1, C_2, C_3, C_4\}$ & $\{N_1, N_2,N_3,N_4\}$ & \textbf{Down. Ratio} & \textbf{Configuration} \\   \midrule
$\{32, 64, 128, 256\}$ & $\{1, 4, 4, 2\}$ & $1/4, 1/8, 1/16, 1/32$ & \{M W W W\} \\ \midrule
\multicolumn{3}{c|}{\textbf{Params and FLOPs$\dagger$ for each stage}} & \textbf{Overall}\\ \midrule
\multicolumn{3}{c|}{0.068M, 0.465M, 1.806M, 3.632M} & \textbf{5.971M}\\ \midrule
\multicolumn{3}{c|}{7.253G, 7.702G, 7.441G, 3.744G} & \textbf{26.14G}\\
\bottomrule
\end{tabular}  }
\caption{Detailed configurations of SPEANet. $C_i$ means the number of output channels in each stage and $N_i$ means the number of HPRBs in each stage. \textbf{Configuration} denotes the strategy adopted at each stage, while `M' and `W' stand for MDP-FASE and WASM. Both the parameters and \textbf{FLOPs} reported for each stage include those of the stage itself and its associated downsampling module. \textbf{Overall} only represents the backbone network, and $\dagger$ indicates that \textbf{Params} and \textbf{FLOPs} are measured using an input $\mathbf{I}\in\mathbb{R}^{1024\times1024\times3}$.
\label{tab:speanet_configuration}
}
\end{table}

\section{Experimental Datasets}
This section provides a detailed overview of the five benchmark datasets employed in evaluation experiments, including remote sensing and multispectral object detection.
\subsubsection{DOTA-v1.0.} DOTA-v1.0~\cite{xia2018dota} is a large-scale benchmark for object detection in aerial images. It contains 2,806 high-resolution images (ranging from 800$\times$800 to 20,000$\times$20,000 pixels) and 188,282 object instances across 15 categories. Objects are annotated with oriented bounding boxes (OBB) using arbitrary quadrilaterals, making it a standard for evaluating oriented object detection algorithms. The 15 object categories include: Plane (PL), Baseball diamond (BD), Bridge (BR), Ground track field (GTF), Small vehicle (SV), Large vehicle (LV), Ship (SH), Tennis court (TC), Basketball court (BC), Storage tank (ST), Soccer-ball field (SBF), Roundabout (RA), Harbor (HA), Swimming pool (SP), and Helicopter (HC).

\subsubsection{DOTA-v1.5.} DOTA-v1.5~\cite{xia2018dota} is an updated version of DOTA-v1.0, using the same images but with revised annotations. It addresses challenges such as small object detection by adding numerous new instances, bringing the total to 403,318. It also introduces a new category, "container crane (CC)," increasing the total to 16.

\subsubsection{DIOR-R.} DIOR-R~\cite{cheng2022anchor} is an extension of the DIOR dataset, tailored for oriented object detection. It comprises 23,463 images and 192,472 instances across 20 object categories, with spatial resolutions from 0.5 to 30 meters. All objects are annotated with OBB, providing a rich resource for developing and testing robust oriented detectors. The 20 categories include: Airplane (APL), Airport (APO), Baseball field (BF), Basketball court (BC), Bridge (BR), Chimney (CH), Expressway service area (ESA), Expressway toll station (ETS), Dam (DAM), Golf field (GF), Ground track field (GTF), Harbor (HA), Overpass (OP), Ship (SH), Stadium (STA), Storage tank (STO), Tennis court (TC), Train station (TS), Vehicle (VE) and Windmill (WM).

\subsubsection{SODA-A.} SODA-A~\cite{cheng2023towards} is a large-scale small object detection dataset that concentrates on aerial scenarios. It contains 2,513 high-resolution images (average resolution is 4,761$\times$2,777) and 872,069 annotations across 9 categories. Objects are annotated with OBB. Categories include: Airplane, Helicopter, Small-vehicle, Large-vehicle, Ship, Container, Storage-tank, Swimming-pool, Windmill.

\subsubsection{MODA.} MODA~\cite{han2026moda} is the first large-scale benchmark for multispectral aerial object detection. It contains 14,041 images with a resolution of 1,200$\times$900 and 330,191 instances across 8 categories. Objects are annotated with OBB. The 8 categories include: Car, Bus, Van, Awning-bike, Truck, Tricycle, Bike, Pedestrian.
\begin{table}[t] \centering	\resizebox{\linewidth}{!}{
\begin{tabular}[c]{l|cccc}
\toprule
\multicolumn{1}{c|}{\textbf{Dataset Name}} & \textbf{Images} & \textbf{Classes} & \textbf{Instances} & \textbf{Resolution} \\
\midrule
DOTA-v1.0 & 2,806 & 15 & 188,282 & $\mathrm{800}^2$-$\mathrm{20,000}^2$ \\
DOTA-v1.5 & 2,806 & 16 & 403,318 & $\mathrm{800}^2$-$\mathrm{20,000}^2$ \\
DIOR-R & 23,463 & 20 & 192,472 & 800$\times$800  \\ 
SODA-A & 2,513 & 9  & 872,069 & 4,761$\times$2,777$^{\dagger}$\\
MODA & 14,041 & 8 & 330,191 & 1,200$\times$900  \\ 
\bottomrule
\end{tabular}  }
\caption {Summary of five benchmark datasets. $^{\dagger}$ represents average resolution.}
\label{tab:dataset_summary}
\end{table}

\section{Experimental Setup}
This section details the experimental protocols, including training configurations, data processing, and evaluation environments for object detection. \\
\textbf{Data Preprocessing:} For DOTA-v1.0 and DOTA-v1.5, original images were cropped into 1,024$\times$1,024 patches with a 200-pixel overlap. For DIOR-R, the input size was 800$\times$800. For SODA-A, original images were cropped into $800\times800$ patches, and the resulting patches were resized into a resolution of $1,200\times1,200$ for training and testing. For MODA dataset, images were resized to $1,200\times1,200$ resolution for both training and testing and the number of input channels was modified to 8 to accommodate multispectral data. \\
\textbf{Training Schedule:} A single-scale training and testing protocol with a $3\times$ training schedule was adopted in experiments. The training process utilized the AdamW optimizer~\cite{loshchilov2017decoupled} with an initial learning rate of $2\times10^{-4}$, betas of $(0.9,0.999)$, and a weight decay of 0.05. The learning rate was decayed by a factor of 0.1 at the 24th and 33rd epochs. A linear warm-up strategy was applied during the first 500 iterations. \\
\textbf{Data Augmentation:} Random flipping along horizontal, vertical and diagonal directions with a per-direction probability of 25\% was adopted, while random rotation was applied to input images with a 50\% probability as well. \\
\textbf{Evaluation:} We followed the standard dataset-specific evaluation protocols. Specifically, for DOTA-v1.0 and DOTA-v1.5, we trained the models on the combined training and validation sets (\textit{trainval}) and evaluated them on the test sets. The detection results were submitted to the official evaluation servers to obtain the final performance scores. For DIOR-R, we utilized both the training and validation sets for model training and evaluated the trained model on the test set. For SODA-A and MODA, the models were trained on the training sets and evaluated on the corresponding test sets. \\
\textbf{Experiment:} The backbone models were pre-trained on ImageNet-1K~\cite{deng2009imagenet} for 300 epochs. All experiments were conducted on the MMRotate framework~\cite{zhou2022mmrotate} and PyTorch~\cite{paszke2019pytorch} on Ubuntu 20.04 with 2 NVIDIA RTX 3090 GPUs.

\begin{table*}[t]    \centering    \setlength{\tabcolsep}{3pt}
\setlength{\HeaderShift}{\dimexpr-\aboverulesep/2-\belowrulesep/2-\cmidrulewidth/2\relax}
\scriptsize
\setlength{\tabcolsep}{2pt}
\resizebox{\linewidth}{!}{
\begin{tabular}[c]{l|c|c|ccccccccccccccc|c}
\toprule
\multicolumn{1}{c|}{\textbf{Method}} & \textbf{Backbone}& \textbf{\#P$\downarrow$} & PL & BD & BR & GTF & SV & LV & SH & TC & BC & ST & SBF & RA & HA & SP & HC &\textbf{mAP$\uparrow$}\\
\midrule
\multicolumn{19}{l}{\textit{One Stage}}\\  \midrule
R$^3$Det~\cite{yang2021r3det} & ResNet-50~\cite{he2016deep} & 41.9M & 89.00 & 75.60 & 46.64 & 67.09 & 76.18 & 73.40 & 79.02 & 90.88 & 78.62 & 84.88 & 59.00 & 61.16 & 63.65 & 62.39 & 37.94 & 69.70\\  \midrule
SASM~\cite{hou2022shape} & ResNet-50~\cite{he2016deep} & 36.6M & 86.42 & 78.97 & 52.47 & 69.84 & 77.30 & 75.99 & 86.72 & \underline{90.89} & 82.63 & 85.66 & 60.13 & 68.25 & 73.98 & 72.22 & 62.37 & 74.92 \\  \midrule
O-RepPoints~\cite{li2022oriented} & ResNet-50~\cite{he2016deep} & 36.6M & 87.02 & 83.17 & 54.13 & 71.16 & 80.18 & 78.40 & 87.28 & \textbf{90.90} & 85.97 & 86.25 & 59.90 & \underline{70.49} & 73.53 & 72.27 & 58.97 & 75.97 \\  \midrule
R$^3$Det-GWD~\cite{yang2021rethinking} & ResNet-50~\cite{he2016deep} & 41.9M & 88.82 & 82.94 & 55.63 & 72.75 & 78.52 & 83.10 & 87.46 & 90.21 & 86.36 & 85.44 & 64.70 & 61.41 & 73.46 & 76.94 & 57.38 & 76.34 \\  \midrule
R$^3$Det-KLD~\cite{yang2021learning} & ResNet-50~\cite{he2016deep} & 41.9M & 88.90 & 84.17 & 55.80 & 69.35 & 78.72 & 84.08 & 87.00 & 89.75 & 84.32 & 85.73 & 64.74 & 61.80 & 76.62 & 78.49 & \textbf{70.89} & 77.36 \\  \midrule
\multirow[c]{4}{*}[\dimexpr 3\HeaderShift\relax]{S$^2$ANet~\cite{han2021align}} & ResNet-50~\cite{he2016deep}  & 38.5M & 89.11 & 82.84 & 48.37 & 71.11 & 78.11 & 78.39 & 87.25 & 90.83 & 84.90 & 85.64 & 60.36 & 62.60 & 65.26 & 69.13 & 57.94 & 74.12 \\ \cmidrule(l){2-19}
& ARC-R50~\cite{pu2023adaptive} & 71.8M & 89.28 & 78.77 & 53.00 & 72.44 & 79.81 & 77.84 & 86.81 & 90.88 & 84.27 & 86.20 & 60.74 & 68.97 & 66.35 & 71.25 & 65.77 & 75.49 \\ \cmidrule(l){2-19}
& PKINet-S~\cite{cai2024poly}  & 24.8M & 89.67 & 84.16 & 51.94 & 71.89 & 80.81 & 83.47 & 88.29 & 90.80 & 87.01 & \textbf{86.94} & 65.02 & 69.53 & 75.83 & \underline{80.20} & 61.85 & 77.83 \\
\cmidrule(l){2-19}
& \textbf{SPEANet} & \textbf{14.0M} & 89.42 & 82.97 & 53.89 & 73.88 & \textbf{82.13} & \underline{85.82} & \underline{88.31} & \underline{90.89} & 86.20 & 86.43 & 62.17 & 67.53 & 77.13 & 80.18 & 65.41 & 78.16 \\ \midrule
\multicolumn{19}{l}{\textit{Two Stage}}\\ \midrule
CenterMap~\cite{CenterMap}  & ResNet-50~\cite{he2016deep}  & 41.1M & 89.02 & 80.56 & 49.41 & 61.98 & 77.99 & 74.19 & 83.74 & 89.44 & 78.01 & 83.52 & 47.64 & 65.93 & 63.68 & 67.07 & 61.59 & 71.59 \\ \midrule
SCRDet~\cite{yang2019scrdet}  & ResNet-50~\cite{he2016deep}  & 41.9M & \textbf{89.98} & 80.65 & 52.09 & 68.36 & 68.36 & 60.32 & 72.41 & 90.85 & \textbf{87.94} & \underline{86.86} & 65.02 & 66.68 & 66.25 & 68.24 & 65.21 & 72.61 \\ \midrule
FR-O~\cite{ren2016faster} & ResNet-50~\cite{he2016deep}  & 41.1M & 89.40 & 81.81 & 47.28 & 67.44 & 73.96 & 73.12 & 85.03 & \textbf{90.90} & 85.15 & 84.90 & 56.60 & 64.77 & 64.70 & 70.28 & 62.22 & 73.17 \\ \midrule
Roi Trans.~\cite{ding2019learning}  & ResNet-50~\cite{he2016deep}  & 55.1M & 89.01 & 77.48 & 51.64 & 72.07 & 74.43 & 77.55 & 87.76 & 90.81 & 79.71 & 85.27 & 58.36 & 64.11 & 76.50 & 71.99 & 54.06 & 74.05 \\ \midrule
G.V.~\cite{xu2020gliding} & ResNet-50~\cite{he2016deep}  & 41.1M & 89.64 & \underline{85.00} & 52.26 & \underline{77.34} & 73.01 & 73.14 & 86.82 & 90.74 & 79.02 & 86.81 & 59.55 & \textbf{70.91} & 72.94 & 70.86 & 57.32 & 75.02 \\ \midrule
ReDet~\cite{han2021redet} & ResNet-50~\cite{he2016deep}  & 31.6M & 88.79 & 82.64 & 53.97 & 74.00 & 78.13 & 84.06 & 88.04 & \underline{90.89} & 87.78 & 85.75 & 61.76 & 60.39 & 75.96 & 68.07 & 63.59 & 76.25 \\ \midrule
\multirow[c]{5}{*}[\dimexpr 4\HeaderShift\relax]{O-RCNN~\cite{xie2021oriented}}  & ResNet-50~\cite{he2016deep}  & 41.1M & 89.46 & 82.12 & 54.78 & 70.86 & 78.93 & 83.00 & 88.20 & \textbf{90.90} & 87.50 & 84.68 & 63.97 & 67.69 & 74.94 & 68.84 & 52.28 & 75.87 \\ \cmidrule(l){2-19}
& ARC-R50~\cite{pu2023adaptive}  & 74.4M & 89.40 & 82.48 & 55.33 & 73.88 & 79.37 & 84.05 & 88.06 & \textbf{90.90} & 86.44 & 84.83 & 63.63 & 70.32 & 74.29 & 71.91 & 65.43 & 77.35 \\
\cmidrule(l){2-19}
& LSKNet-S~\cite{Li_2023_ICCV}  & 31.0M & 89.66 & \textbf{85.52} & \textbf{57.72} & 75.70 & 74.95 & 78.69 & 88.24 & 90.88 & 86.79 & 86.38 & \textbf{66.92} & 63.77 & \underline{77.77} & 74.47 & 64.82 & 77.49 \\ \cmidrule(l){2-19}
& PKINet-S~\cite{cai2024poly}  & 30.8M & \underline{89.72} & 84.20 & 55.81 & \textbf{77.63} & 80.25 & 84.45 & 88.12 & 90.88 & 87.57 & 86.07 & \underline{66.86} & 70.23 & 77.47 & 73.62 & 62.94 & \underline{78.39} \\ \cmidrule(l){2-19}
& \textbf{SPEANet} & \underline{23.0M} & 88.69 & 83.40 & \underline{56.19} & 74.03 & \underline{80.92} & \textbf{85.86} & \textbf{88.46} & 90.68 & \underline{87.81} & 85.66 & 63.32 & 64.29 & \textbf{77.94} & \textbf{80.96} & \underline{70.03} & \textbf{78.55} \\ \bottomrule
\end{tabular}  }
\caption{Performance comparison on the DOTA-v1.0 test set. The best results are shown in bold, and the second-best results are underlined. Results for competing methods follow the corresponding published reports~\cite{lu2025legnet}.}  \label{tab:result_dota1.0_detailed}
\end{table*}

\section{Design and Implementation Details}
This section provides implementation-level details of the fixed structural operators used in SPEANet.

Let $\Omega_k=\{-r,\ldots,r\}^2$ denote the support of an odd-sized
$k\times k$ kernel, where $r=(k-1)/2$. For $(i,j)\in\Omega_k$, the
discrete Gaussian kernel is normalized over its finite support:
\begin{equation}
	K_{\mathrm{G}}(i,j;\sigma)=
	\frac{
		\exp\left(-\frac{i^2+j^2}{2\sigma^2}\right)
	}{
		\sum_{(u,v)\in\Omega_k}
		\exp\left(-\frac{u^2+v^2}{2\sigma^2}\right)
	}.
\end{equation}
Here, $\sigma$ controls the standard deviation of the Gaussian envelope.
The discrete LoG kernel is constructed from an analytical response:
\begin{equation}
	\widetilde{K}_{\mathrm{LoG}}(i,j;\sigma)=
	\frac{i^2+j^2-2\sigma^2}{\sigma^4}
	\exp\left(-\frac{i^2+j^2}{2\sigma^2}\right),
\end{equation}
followed by a zero-mean correction over its finite support:
\begin{equation}
	K_{\mathrm{LoG}}(i,j;\sigma)=
	\widetilde{K}_{\mathrm{LoG}}(i,j;\sigma)
	-\frac{1}{|\Omega_k|}
	\sum_{(u,v)\in\Omega_k}
	\widetilde{K}_{\mathrm{LoG}}(u,v;\sigma).
\end{equation}
This correction suppresses the DC response introduced by truncating the
analytical LoG to a finite discrete kernel.

For first-order directional responses, we explicitly define four fixed
$3\times3$ Sobel kernels corresponding to the horizontal, vertical, and
two diagonal directions:
\[
\mathrm{Sobel}_{x} =
\begin{bmatrix}
	-1 & 0 & 1 \\
	-2 & 0 & 2 \\
	-1 & 0 & 1
\end{bmatrix},
\mathrm{Sobel}_{y} =
\begin{bmatrix}
	-1 & -2 & -1 \\
	0 & 0 & 0 \\
	1 & 2 & 1
\end{bmatrix},
\]
\[
\mathrm{Sobel}_{45^{\circ}} =
\begin{bmatrix}
	-2 & -1 & 0 \\
	-1 & 0 & 1 \\
	0 & 1 & 2
\end{bmatrix},
\mathrm{Sobel}_{135^{\circ}} =
\begin{bmatrix}
	0 & 1 & 2 \\
	-1 & 0 & 1 \\
	-2 & -1 & 0
\end{bmatrix}.
\]
The isotropic second-order response is obtained using the following
eight-neighborhood Laplacian kernel:
\[
\mathrm{Laplacian} =
\begin{bmatrix}
	1 & 1 & 1 \\
	1 & -8 & 1 \\
	1 & 1 & 1
\end{bmatrix}.
\]

These fixed operators are applied to multi-channel feature maps through
grouped depthwise convolution. By setting
\texttt{groups=channels}, the same fixed operator or operator bank is
applied independently to every input channel, without cross-channel
interaction at the fixed-filtering stage. Although the kernels are
instantiated as convolutional weights, they are frozen by setting
\texttt{requires\_grad=False}; therefore, they introduce no trainable
degrees of freedom and receive no gradient updates during optimization.

To further reduce the parameter overhead, the context-gate generator of FASE module employs $1\times1$ pointwise convolutions for channel projection and dimensionality reduction, together with depthwise convolutions for parameter-efficient spatial feature extraction. These operations jointly contribute to the compact parameter footprint of SPEANet. In WASM, both the approximation and detail branches are implemented using depthwise separable convolutions, each consisting of a $3\times3$ depthwise convolution followed by a $1\times1$ pointwise convolution. This design decouples spatial modeling from cross-channel interaction, thereby reducing both the number of parameters and the computational cost.

In HPRB, the internal FASE module already incorporates normalization and a residual connection. Therefore, the effective residual branch is implemented as \texttt{x\_out - x}. DropPath is then applied only to this residual component, and the resulting feature is added back to the original input. Formally, this operation is expressed as
\[
\mathbf{Y}=\mathbf{X}+\operatorname{DropPath}\left(\mathbf{X}_{\mathrm{out}}-\mathbf{X}\right).
\]

\begin{table*}[t]	\centering	    \scriptsize
\setlength{\tabcolsep}{2pt}
\resizebox{\textwidth}{!}{
\begin{tabular}[c]{l|c|cccccccccccccccc|c}
\toprule
\multicolumn{1}{c|}{\textbf{Method}} & \textbf{\#P$\downarrow$} & PL & BD & BR & GTF & SV & LV & SH & TC & BC & ST & SBF & RA & HA & SP & HC & CC & \textbf{mAP$\uparrow$}\\
\midrule
RetinaNet-O~\cite{lin2017focal} & 36.4M & 71.43 & 77.64 & 42.12 & 64.65 & 44.53 & 56.79 & 73.31 & 90.84 & 76.02 & 59.96 & 46.95 & 69.24 & 59.65 & 64.52 & 48.06 & 0.83 & 59.16 \\
\midrule
FR-O~\cite{ren2016faster} & 41.1M &  71.89 & 74.47 & 44.45 & 59.87 & 51.28 & 68.98 & 79.37 & 90.78 & 77.38 & 67.50 & 47.75 & 69.72 & 61.22 & 65.28 & 60.47 & 1.54 & 62.00 \\
\midrule
Mask R-CNN~\cite{he2017piotr} & 44.4M & 76.84 & 73.51 & 49.90 & 57.80 & 51.31 & 71.34 & 79.75 & 90.46 & 74.21 & 66.07 & 46.21 & 70.61 & 63.07 & 64.46 & 57.81 & 9.42 & 62.67 \\
\midrule
HTC~\cite{chen2019hybrid} & 77.5M & 77.80 & 73.67 & 51.40 & 63.99 & 51.54 & 73.31 & 80.31 & 90.48 & 75.12 & 67.34 & 48.51 & 70.63 & 64.84 & 64.48 & 55.87 & 5.15 & 63.40 \\
\midrule
ReDet~\cite{han2021redet} & 31.6M & 79.20 & 82.81 & 51.92 & 71.41 & 52.38 & 75.73 & 80.92 & 90.83 & 75.81 & 68.64 & 49.29 & 72.03 & 73.36 & 70.55 & 63.33 & 11.53 & 66.86 \\
\midrule
LSKNet-S~\cite{Li_2023_ICCV}& 31.0M & 72.05 & \underline{84.94} & \underline{55.41} & \textbf{74.93} & 52.42 & \underline{77.45} & 81.17 & \underline{90.85} & \underline{79.44} & 69.00 & 62.10 & \textbf{73.72} & \textbf{77.49} & \underline{75.29} & 55.81 & \textbf{42.19} & 70.26 \\
\midrule
SOOD~\cite{xi2024structure} & - & \textbf{80.32} & 84.41 & 52.59 & \underline{74.77} & \textbf{58.48} & 76.90 & 86.97 & \textbf{90.87} & 78.62 & \textbf{76.56} & \underline{62.93} & 71.16 & 74.64 & \textbf{76.04} & 55.97 & 25.09 & 70.39 \\
\midrule
SPCNet~\cite{zheng2025spcnet}& 25.3M & 79.82 & 84.25 & 53.78 & 74.24 & 52.15 & 76.66 & 87.35 & \underline{90.85} & 78.65 & \underline{69.45} & 61.95 & \underline{72.98} & 76.11 & 72.96 & 61.25 & 37.18 & 70.60 \\
\midrule
PKINet-S~\cite{cai2024poly} & 30.8M & \underline{80.31} & \textbf{85.00} & \textbf{55.61} & 74.38 & 52.41 & 76.85 & \underline{88.38} & \textbf{90.87} & 79.04 & 68.78 & \textbf{67.47} & 72.45 & \underline{76.24} & 74.53 & \underline{64.07} & 37.13 & \underline{71.47} \\
\midrule
\rowcolor{gray!25}
\textbf{SPEANet} & \textbf{23.0M} & 80.13 & 83.23 & 55.03 & 72.60 & \underline{52.69} & \textbf{82.00} & \textbf{88.61} & 90.67 & \textbf{83.42} & 68.79 & 61.95 & 72.83 & 76.15 & 72.74 & \textbf{74.75} & \underline{40.25} & \textbf{72.24} \\
\bottomrule
\end{tabular}
}
\caption {Performance comparison on DOTA-v1.5 test set. The best results are shown in bold and the second-best results are underlined. Results for competing methods follow the corresponding published reports~\cite{lu2025legnet,wu2025measuring}.}
\label{tab:result_dota15_detailed}
\end{table*}

\section{Quantitative Results}
To complement the aggregate results reported in the main paper, Tables~\ref{tab:result_dota1.0_detailed} and~\ref{tab:result_dota15_detailed} provide the complete category-wise AP results on DOTA-v1.0 and DOTA-v1.5, respectively.

As shown in Table~\ref{tab:result_dota1.0_detailed}, O-RCNN with SPEANet achieves the best or second-best AP on 8 of the 15 categories. Relative to the strongest non-SPEANet competitors, it improves LV and SP by 1.41 and 0.76 AP points. More importantly, the gains are consistent across detectors: the SPEANet variants under S$^2$ANet and O-RCNN occupy the top two positions on SV, LV, and SH. This detector-consistent behavior suggests that the improvements are associated with the backbone representation rather than a particular detection head.

Under the controlled S$^2$ANet, SPEANet improves SV and LV by 1.32 and 2.35 AP points over PKINet-S, respectively, while using 14.0M rather than 24.8M parameters. It also improves mAP by
4.04 points over ResNet-50 with substantially fewer parameters. The gains on boundary- and scale-sensitive categories are consistent with the intended role of structural priors in preserving localized contour evidence. Nevertheless, this category-level evidence does not isolate the contribution of individual priors, and near-ties on saturated categories such as TC should not be overinterpreted.

Table~\ref{tab:result_dota15_detailed} reports the complete category-wise results on DOTA-v1.5. SPEANet achieves the highest overall mAP of 72.24 with 23.0M parameters, exceeding PKINet-S by
0.77 AP points while using 25.3\% fewer reported parameters. It ranks first on LV, SH, BC, and HC and second on SV and CC. The advantages are particularly pronounced on LV, BC, and HC, where SPEANet surpasses the corresponding second-best results by 4.55, 3.98, and 10.68 AP points, respectively. Together with the strong LV, SH, and HC results on DOTA-v1.0, this cross-version consistency
suggests that the learned representation transfers reliably to these categories rather than benefiting only a particular dataset split.

The improvements are nevertheless category-dependent. SPEANet does not lead on GTF, ST, SBF, or SP, and its second-ranked SV result remains below SOOD by 5.79 AP points. Therefore, the category-wise
results are consistent with the intended benefit of structural priors for boundary- and localization-sensitive objects, but do not by themselves establish a causal contribution from individual Sobel or Laplacian cues. Overall, SPEANet provides a favorable accuracy--parameter trade-off rather than uniform superiority across all categories.

\begin{table*}[ht]
    \centering
    \scriptsize
    \setlength{\tabcolsep}{2pt}
    \resizebox{\linewidth}{!}{
        \begin{tabular}{cccccc|c|*{15}{c}|c}
            \toprule
            \multicolumn{2}{c}{CLoG-Stem} & \multicolumn{2}{c}{MDP-FASE} & \multicolumn{2}{c|}{WASM} & \textbf{\#P$\downarrow$} & PL & BD & BR & GTF & SV & LV & SH & TC & BC & ST & SBF & RA & HA & SP & HC & \textbf{mAP$\uparrow$} \\
            \midrule
            \multicolumn{2}{c}{\ding{51}} & \multicolumn{2}{c}{} & \multicolumn{2}{c|}{} & 5.351M & \textbf{89.23} & 82.10 & 54.15 & 73.79 & 78.75 & 84.51 & 87.89 & \textbf{90.73} & 87.09 & \underline{85.94} & 61.44 & 62.98 & 75.60 & 72.81 & 61.70 & 76.581 \\
            \multicolumn{2}{c}{} & \multicolumn{2}{c}{\ding{51}} & \multicolumn{2}{c|}{} & 6.526M & 88.85 & 82.19 & \underline{55.88} & 72.27 & 78.43 & \underline{85.60} & 88.30 & 90.67 & 85.70 & 85.23 & 63.70 & 65.03 & 76.83 & 73.23 & 65.86 & 77.184 \\
            \multicolumn{2}{c}{} & \multicolumn{2}{c}{} & \multicolumn{2}{c|}{\ding{51}} & \underline{4.786M} & 88.97 & 83.27 & 53.57 & \textbf{76.69} & 78.53 & 84.53 & 87.88 & 90.62 & 84.77 & 84.40 & 63.92 & 65.12 & 76.98 & 78.62 & 66.20 & 77.605 \\
            \multicolumn{2}{c}{\ding{51}} & \multicolumn{2}{c}{\ding{51}} & \multicolumn{2}{c|}{} & 6.531M & 89.02 & 83.10 & 55.23 & 73.49 & 79.38 & 85.15 & 88.36 & \textbf{90.73} & 87.15 & 85.06 & \textbf{66.90} & \textbf{66.86} & 76.84 & \underline{80.24} & 63.30 & \underline{78.054} \\
            \multicolumn{2}{c}{\ding{51}} & \multicolumn{2}{c}{} & \multicolumn{2}{c|}{\ding{51}} & 4.791M & 88.90 & \textbf{83.47} & 54.92 & \underline{75.76} & 78.86 & 84.92 & 88.17 & \underline{90.72} & 87.40 & 84.75 & 63.25 & 63.33 & 76.92 & 78.99 & 66.29 & 77.778 \\
            \multicolumn{2}{c}{} & \multicolumn{2}{c}{\ding{51}} & \multicolumn{2}{c|}{\ding{51}} & 5.966M & 89.07 & 82.94 & 55.55 & 73.59 & \underline{79.96} & 85.51 & 88.16 & 90.55 & \textbf{87.98} & \textbf{86.22} & \underline{64.86} & \underline{66.85} & \underline{77.09} & 73.52 & \underline{68.47} & 78.022 \\
            \midrule
            \multicolumn{6}{@{}l@{}|}{\hfill\makebox[0pt][c]{CGFF}\hfill\hfill\makebox[0pt][c]{FASE}\hfill} & & & & & & & & & & & & & & & & & \\
            \midrule
            \multicolumn{6}{@{}l@{}|}{\hfill\makebox[0pt][c]{\ding{51}}\hfill\hfill\makebox[0pt][c]{}\hfill} & \textbf{4.681M} & \underline{89.11} & 83.26 & 55.20 & 73.25 & 78.65 & 84.80 & \textbf{88.53} & 90.46 & 86.45 & 85.72 & 63.26 & 63.22 & 76.52 & 79.90 & 64.50 & 77.521 \\
            \multicolumn{6}{@{}l@{}|}{\hfill\makebox[0pt][c]{}\hfill\hfill\makebox[0pt][c]{\ding{51}}\hfill} & 5.019M & 88.64 & 82.97 & 54.62 & 74.14 & 78.87 & 84.60 & 88.17 & 90.56 & 86.15 & 85.43 & 62.10 & 58.82 & 76.47 & 73.71 & 68.07 & 76.887 \\
            \midrule
            \rowcolor{gray!25}
            \multicolumn{6}{c|}{\textbf{SPEANet}} & 5.971M & 88.69 & \underline{83.40} & \textbf{56.19} & 74.03 & \textbf{80.92} & \textbf{85.86} & \underline{88.46} & 90.68 & \underline{87.81} & 85.66 & 63.32 & 64.29 & \textbf{77.94} & \textbf{80.96} & \textbf{70.03} & \textbf{78.550} \\
            \bottomrule
        \end{tabular}
    }
    \caption{Ablation of SPEANet on DOTA-v1.0. All variants use 100-epoch ImageNet-1K pretraining and the Oriented R-CNN detector. \#P denotes backbone parameters.}
    \label{tab:ablation_detailed}
\end{table*}

\begin{figure*}[!t] \centering
	\includegraphics[width=1.0\textwidth]{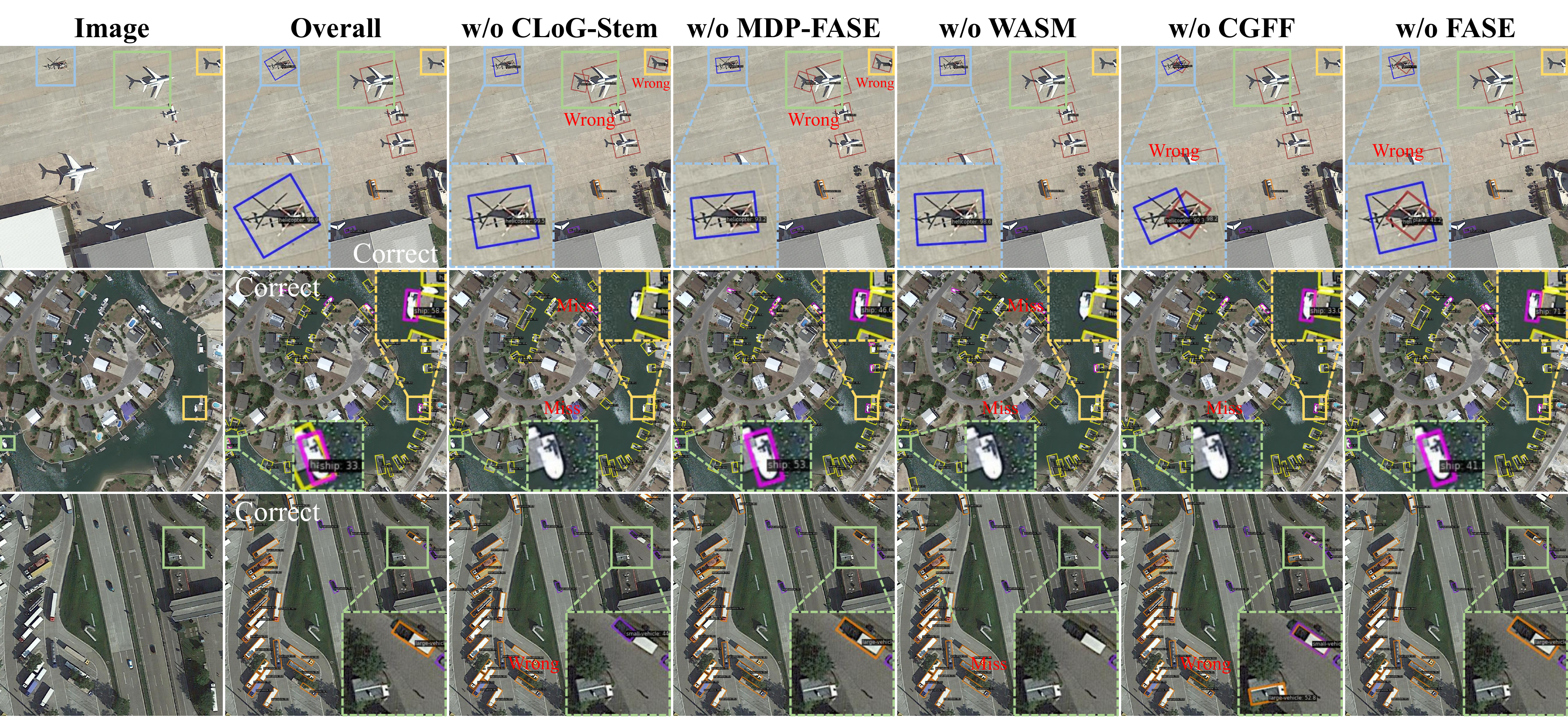}
	\caption{Visual comparison of different SPEANet variants in the ablation study. ``Overall'' represents the complete SPEANet model while ``w/o'' denotes the removal of the corresponding component.}
	\label{fig_abl_vis}  
\end{figure*}

\section{Ablation Study}

We evaluate the three stage-level components, CLoG-Stem, MDP-FASE, and WASM, together with the internal CGFF and FASE mechanisms. When a component is disabled, CLoG-Stem is replaced by a $4\times4$ convolution with stride 4, MDP-FASE and WASM by a $3\times3$ convolution, CGFF by a $3\times3$ convolution, and FASE by a $1\times1$ convolution. These replacements preserve the input and output tensor dimensions; their different capacities are reported explicitly through the backbone parameter count. All variants use the same 100-epoch ImageNet-1K pretraining and Oriented R-CNN evaluation protocol.

Table~\ref{tab:ablation_detailed} shows positive marginal gains from all three stage-level components in every matched configuration, although the magnitude depends on the accompanying modules. CLoG-Stem improves mAP by 0.173--0.870 AP points with only 0.005M additional parameters. MDP-FASE provides gains of 0.417--1.473 AP points at a cost of 1.180M parameters. WASM improves mAP by 0.496--1.197 AP points while reducing the parameter count by 0.560M in the corresponding comparisons. Among the three single-component configurations, WASM obtains the highest mAP of 77.605 with 4.786M parameters, supporting its use as a parameter-efficient operator in the deeper stages.

The CGFF and FASE results further indicate complementary effects. Adding CGFF to the FASE-only configuration increases mAP from 76.887 to 78.550, whereas adding FASE to the CGFF-only configuration raises it from 77.521 to 78.550. These improvements correspond to gains of 1.663 and 1.029 AP points. Thus, neither mechanism alone reproduces the performance of their joint configuration.

The complete SPEANet ranks first or second on 9 of the 15 categories. It obtains the highest AP on BR, SV, LV, HA, SP, and HC, exceeding the corresponding second-best variants by 0.31, 0.96, 0.26, 0.85, 0.72, and 1.56 AP points, respectively. It also ranks second on BD, SH, and BC. These results indicate that the complete configuration favors balanced category-level performance rather than optimizing a single object type.

We visualize the detection results of different SPEANet variants in the ablation study. As shown in Figure~\ref{fig_abl_vis}, removing any individual component leads to missed detections or false detections. These qualitative results further illustrate the contribution of each component and the effectiveness of their collaborative integration in the complete SPEANet.

\begin{figure*}[t]
	\centering
	\includegraphics[width=\textwidth]{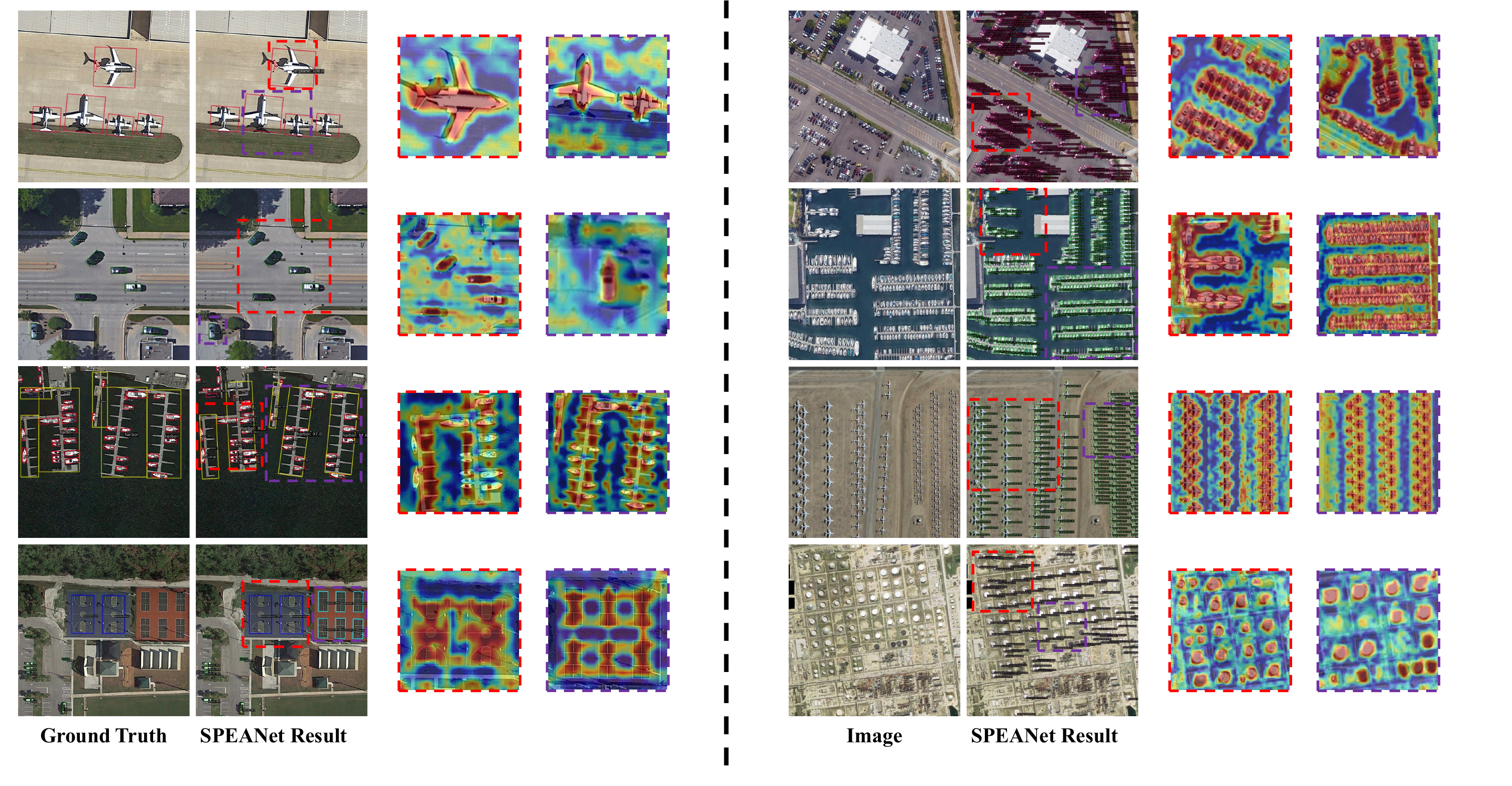}
	\vspace{2mm}
	\includegraphics[width=\textwidth]{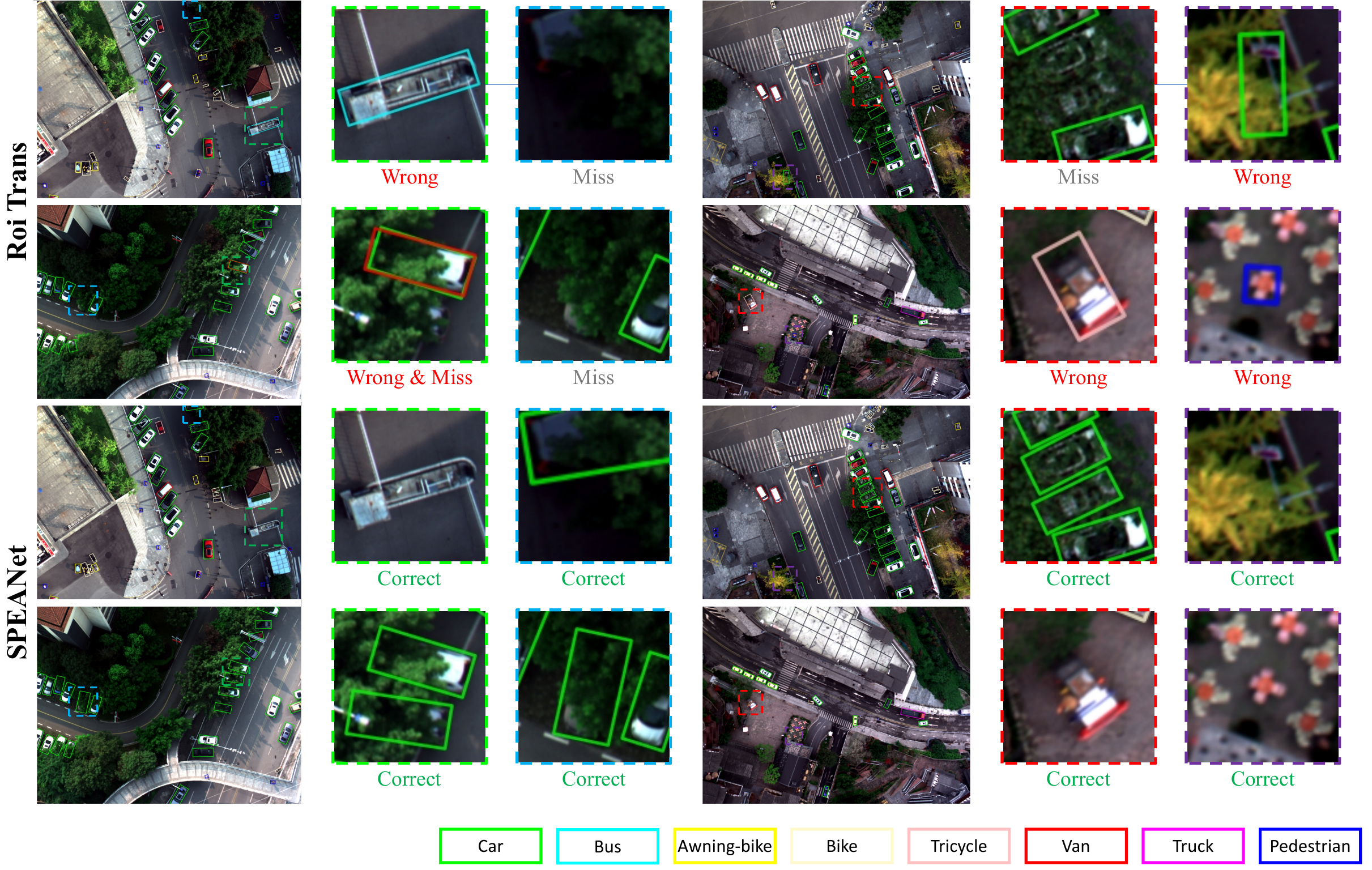}
	\caption{Qualitative visualization results of SPEANet. \textbf{Top:} Visualization of detection results and CGFF gating maps on different datasets. The left and right examples are from DIOR-R and SODA-A, respectively. \textbf{Bottom:} Visualization of detection results compared with Roi Trans.~\cite{ding2019learning} on MODA dataset.}
	\label{fig:qualitative_results}
\end{figure*}

\section{Qualitative Results}
\begin{figure*}[t]
    \centering
    \includegraphics[width=0.91\textwidth]{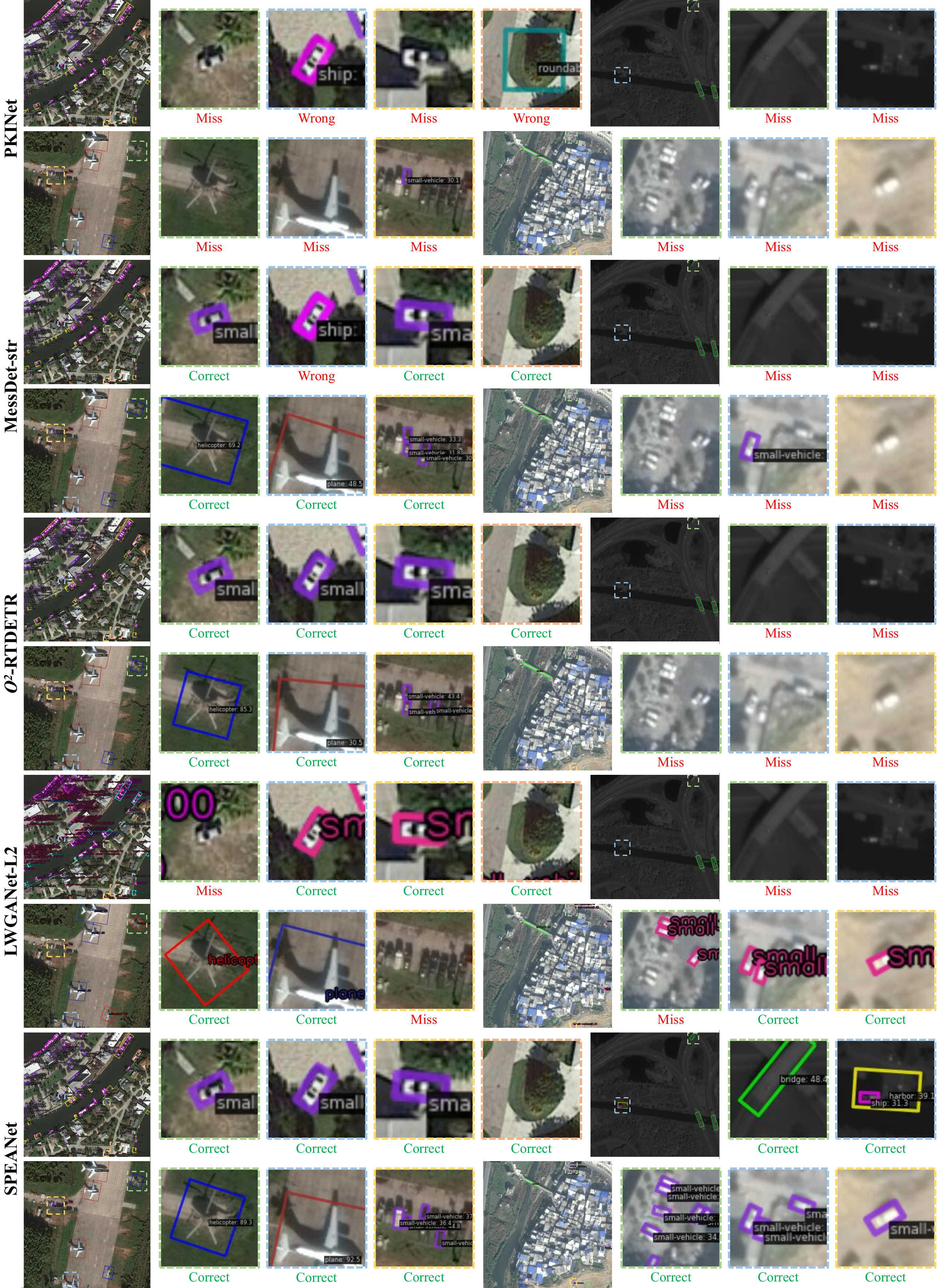}
    \caption{Comparison of the detection results obtained by different methods on the DOTA-v1.5 test set.}
    \label{fig_vis5}
\end{figure*}

As shown in Figure~\ref{fig:qualitative_results}, the visualization results on the DIOR-R dataset demonstrate that SPEANet can accurately localize objects across diverse and complex remote-sensing scenarios, highlighting its robustness and generalization capability under varying scene conditions. On the SODA-A dataset, SPEANet detects more small objects with fewer missed detections in the selected examples. Furthermore, the gating maps generated by the CGFF module consistently highlight the spatial regions corresponding to these objects, supporting the effectiveness of CGFF module in enhancing small-object feature representations. For visualization purposes, we select three of the eight channels from each MODA image to generate the displayed images. As shown in Figure~\ref{fig:qualitative_results}, the qualitative results show that SPEANet remains effective in detecting partially occluded objects, further demonstrating its robustness in multispectral imaging scenarios. As shown in Figure~\ref{fig_vis5}, similar advantages can also be observed on the DOTA-v1.5 dataset, where SPEANet achieves more reliable detection results than PKINet~\cite{cai2024poly}, MessDet~\cite{wu2025measuring}, $O^2$-RTDETR~\cite{ding2026real}, and LWGANet~\cite{lu2026lwganet}. In particular, for commonly occurring categories such as small vehicle and ship, SPEANet produces neither missed detections nor false positives in the illustrated cases.

\bibliography{speanet}

@String(AAAI  = {Proc. AAAI Conf. Artif. Intell.})

@String(ACMMM = {Proc. ACM Int. Conf. Multimedia})

@String(CVPR  = {Proc. IEEE Conf. Comput. Vis. Pattern Recog.})

@String(ECCV  = {Proc. Eur. Conf. Comput. Vis.})

@String(ICCV  = {Proc. IEEE Int. Conf. Comput. Vis.})

@String(ICCVW = {Proc. IEEE Int. Conf. Comput. Vis. Worksh.})

@String(ICLR  = {Proc. Int. Conf. Learn. Represent.})

@String(ICML  = {Proc. Int. Conf. Mach. Learn.})

@String(NIPS  = {Proc. Adv. Neural Inform. Process. Syst.})

@String(ISPRS  = {ISPRS J. Photogramm. Remote Sens.})

@String(RS     = {Remote Sens.})

@String(TGRS   = {IEEE Trans. Geosci. Remote Sens.})

@String(TIP    = {IEEE Trans. Image Process.})

@String(TPAMI  = {IEEE Trans. Pattern Anal. Mach. Intell.})

@inproceedings{cai2024poly,
	author    = {Cai, Xinhao and Lai, Qiuxia and Wang, Yuwei and Wang, Wenguan and Sun, Zeren and Yao, Yazhou},
	title     = {Poly Kernel Inception Network for Remote Sensing Detection},
	booktitle = CVPR,
	pages     = {27706--27716},
	year      = {2024}
}

@inproceedings{chen2019hybrid,
	author    = {Chen, Kai and Pang, Jiangmiao and Wang, Jiaqi and Xiong, Yu and Li, Xiaoxiao and Sun, Shuyang and Feng, Wansen and Liu, Ziwei and Shi, Jianping and Ouyang, Wanli and Loy, Chen Change and Lin, Dahua},
	title     = {Hybrid Task Cascade for Instance Segmentation},
	booktitle = CVPR,
	pages     = {4974--4983},
	year      = {2019}
}

@inproceedings{deng2009imagenet,
	author    = {Deng, Jia and Dong, Wei and Socher, Richard and Li, Li-Jia and Li, Kai and Fei-Fei, Li},
	title     = {{ImageNet}: A Large-Scale Hierarchical Image Database},
	booktitle = CVPR,
	pages     = {248--255},
	year      = {2009}
}

@inproceedings{ding2019learning,
	author    = {Ding, Jian and Xue, Nan and Long, Yang and Xia, Gui-Song and Lu, Qikai},
	title     = {Learning {RoI} Transformer for Oriented Object Detection in Aerial Images},
	booktitle = CVPR,
	pages     = {2849--2858},
	year      = {2019}
}

@inproceedings{guo2021beyond,
	author    = {Guo, Zonghao and Liu, Chang and Zhang, Xiaosong and Jiao, Jianbin and Ji, Xiangyang and Ye, Qixiang},
	title     = {Beyond Bounding-Box: Convex-Hull Feature Adaptation for Oriented and Densely Packed Object Detection},
	booktitle = CVPR,
	pages     = {8792--8801},
	year      = {2021}
}

@inproceedings{han2026moda,
	author    = {Han, Shuaihao and Xu, Tingfa and Liu, Peifu and Li, Jianan},
	title     = {{MODA}: The First Challenging Benchmark for Multispectral Object Detection in Aerial Images},
	booktitle = AAAI,
	volume    = {40},
	pages     = {4574--4582},
	year      = {2026}
}

@inproceedings{han2021redet,
	author    = {Han, Jiaming and Ding, Jian and Xue, Nan and Xia, Gui-Song},
	title     = {{ReDet}: A Rotation-Equivariant Detector for Aerial Object Detection},
	booktitle = CVPR,
	pages     = {2786--2795},
	year      = {2021}
}

@inproceedings{he2016deep,
	author    = {He, Kaiming and Zhang, Xiangyu and Ren, Shaoqing and Sun, Jian},
	title     = {Deep Residual Learning for Image Recognition},
	booktitle = CVPR,
	pages     = {770--778},
	year      = {2016}
}

@inproceedings{he2017piotr,
	author    = {He, Kaiming and Gkioxari, Georgia and Doll{\'a}r, Piotr and Girshick, Ross},
	title     = {Mask {R-CNN}},
	booktitle = ICCV,
	pages     = {2961--2969},
	year      = {2017}
}

@inproceedings{hou2022shape,
	author    = {Hou, Liping and Lu, Ke and Xue, Jian and Li, Yuqiu},
	title     = {Shape-Adaptive Selection and Measurement for Oriented Object Detection},
	booktitle = AAAI,
	volume    = {36},
	pages     = {923--932},
	year      = {2022}
}

@inproceedings{Li_2023_ICCV,
	author    = {Li, Yuxuan and Hou, Qibin and Zheng, Zhaohui and Cheng, Ming-Ming and Yang, Jian and Li, Xiang},
	title     = {Large Selective Kernel Network for Remote Sensing Object Detection},
	booktitle = ICCV,
	pages     = {16794--16805},
	year      = {2023}
}

@inproceedings{li2022oriented,
	author    = {Li, Wentong and Chen, Yijie and Hu, Kaixuan and Zhu, Jianke},
	title     = {Oriented {RepPoints} for Aerial Object Detection},
	booktitle = CVPR,
	pages     = {1829--1838},
	year      = {2022}
}

@inproceedings{lin2017focal,
	author    = {Lin, Tsung-Yi and Goyal, Priya and Girshick, Ross and He, Kaiming and Doll{\'a}r, Piotr},
	title     = {Focal Loss for Dense Object Detection},
	booktitle = ICCV,
	pages     = {2980--2988},
	year      = {2017}
}

@inproceedings{loshchilov2017decoupled,
	author    = {Loshchilov, Ilya and Hutter, Frank},
	title     = {Decoupled Weight Decay Regularization},
	booktitle = ICLR,
	year      = {2019}
}

@article{ding2026real,
  author={Ding, Zeyu and Zhou, Yong and Zhao, Jiaqi and Du, Wen-Liang and Li, Xixi and Yao, Rui and El Saddik, Abdulmotaleb},
  title={Real-time oriented object detection transformer in remote sensing images},
  journal=TGRS,
  year={2026},
}

@inproceedings{lu2025legnet,
	author    = {Lu, Wei and Chen, Si-Bao and Li, Hui-Dong and Shu, Qing-Ling and Ding, Chris H. Q. and Tang, Jin and Luo, Bin},
	title     = {{LEGNet}: A Lightweight Edge-Gaussian Network for Low-Quality Remote Sensing Image Object Detection},
	booktitle = ICCVW,
	pages     = {2844--2853},
	year      = {2025}
}

@inproceedings{lu2026lwganet,
	author    = {Lu, Wei and Yang, Xue and Chen, Si-Bao},
	title     = {{LWGANet}: Addressing Spatial and Channel Redundancy in Remote Sensing Visual Tasks with Light-Weight Grouped Attention},
	booktitle = AAAI,
	volume    = {40},
	pages     = {7574--7582},
	year      = {2026}
}

@inproceedings{paszke2019pytorch,
	author    = {Paszke, Adam and Gross, Sam and Massa, Francisco and Lerer, Adam and Bradbury, James and Chanan, Gregory and Killeen, Trevor and Lin, Zeming and Gimelshein, Natalia and Antiga, Luca and Desmaison, Alban and K{\"o}pf, Andreas and Yang, Edward and DeVito, Zachary and Raison, Martin and Tejani, Alykhan and Chilamkurthy, Sasank and Steiner, Benoit and Fang, Lu and Bai, Junjie and Chintala, Soumith},
	title     = {{PyTorch}: An Imperative Style, High-Performance Deep Learning Library},
	booktitle = NIPS,
	volume    = {32},
	pages     = {8024--8035},
	year      = {2019}
}

@inproceedings{pu2023adaptive,
	author    = {Pu, Yifan and Wang, Yiru and Xia, Zhuofan and Han, Yizeng and Wang, Yulin and Gan, Weihao and Wang, Zidong and Song, Shiji and Huang, Gao},
	title     = {Adaptive Rotated Convolution for Rotated Object Detection},
	booktitle = CVPR,
	pages     = {6589--6600},
	year      = {2023}
}

@inproceedings{wu2025measuring,
	author    = {Wu, Xiuyu and Wang, Xinhao and Zhu, Xiubin and Yang, Lan and Liu, Jiyuan and Hu, Xingchen},
	title     = {Measuring the Impact of Rotation Equivariance on Aerial Object Detection},
	booktitle = ICCV,
	pages     = {7329--7339},
	year      = {2025}
}

@inproceedings{xia2018dota,
	author    = {Xia, Gui-Song and Bai, Xiang and Ding, Jian and Zhu, Zhen and Belongie, Serge and Luo, Jiebo and Datcu, Mihai and Pelillo, Marcello and Zhang, Liangpei},
	title     = {{DOTA}: A Large-Scale Dataset for Object Detection in Aerial Images},
	booktitle = CVPR,
	pages     = {3974--3983},
	year      = {2018}
}

@inproceedings{xie2021oriented,
	author    = {Xie, Xingxing and Cheng, Gong and Wang, Jiabao and Yao, Xiwen and Han, Junwei},
	title     = {Oriented {R-CNN} for Object Detection},
	booktitle = ICCV,
	pages     = {3520--3529},
	year      = {2021}
}

@inproceedings{xu2023dynamic,
	author    = {Xu, Chang and Ding, Jian and Wang, Jinwang and Yang, Wen and Yu, Huai and Yu, Lei and Xia, Gui-Song},
	title     = {Dynamic Coarse-to-Fine Learning for Oriented Tiny Object Detection},
	booktitle = CVPR,
	pages     = {7318--7328},
	year      = {2023}
}

@inproceedings{yang2019scrdet,
	author    = {Yang, Xue and Yang, Jirui and Yan, Junchi and Zhang, Yue and Zhang, Tengfei and Guo, Zhi and Sun, Xian and Fu, Kun},
	title     = {{SCRDet}: Towards More Robust Detection for Small, Cluttered and Rotated Objects},
	booktitle = ICCV,
	pages     = {8232--8241},
	year      = {2019}
}

@inproceedings{yang2020arbitrary,
	author    = {Yang, Xue and Yan, Junchi},
	title     = {Arbitrary-Oriented Object Detection with Circular Smooth Label},
	booktitle = ECCV,
	pages     = {677--694},
	year      = {2020}
}

@inproceedings{yang2021learning,
	author    = {Yang, Xue and Yang, Xiaojiang and Yang, Jirui and Ming, Qi and Wang, Wentao and Tian, Qi and Yan, Junchi},
	title     = {Learning High-Precision Bounding Box for Rotated Object Detection via {Kullback-Leibler} Divergence},
	booktitle = NIPS,
	volume    = {34},
	pages     = {18381--18394},
	year      = {2021}
}

@inproceedings{yang2021r3det,
	author    = {Yang, Xue and Yan, Junchi and Feng, Ziming and He, Tao},
	title     = {{$R^3$Det}: Refined Single-Stage Detector with Feature Refinement for Rotating Object},
	booktitle = AAAI,
	volume    = {35},
	pages     = {3163--3171},
	year      = {2021}
}

@inproceedings{yang2021rethinking,
	author    = {Yang, Xue and Yan, Junchi and Ming, Qi and Wang, Wentao and Zhang, Xiaopeng and Tian, Qi},
	title     = {Rethinking Rotated Object Detection with {Gaussian Wasserstein} Distance Loss},
	booktitle = ICML,
	volume    = {139},
	pages     = {11830--11841},
	year      = {2021}
}

@inproceedings{yuan2026strip,
	author    = {Yuan, Xinbin and Zheng, Zhaohui and Li, Yuxuan and Liu, Xialei and Liu, Li and Li, Xiang and Hou, Qibin and Cheng, Ming-Ming},
	title     = {Strip {R-CNN}: Large Strip Convolution for Remote Sensing Object Detection},
	booktitle = AAAI,
	volume    = {40},
	pages     = {12259--12267},
	year      = {2026}
}

@inproceedings{zhou2022mmrotate,
	author    = {Zhou, Yue and Yang, Xue and Zhang, Gefan and Wang, Jiabao and Liu, Yanyi and Hou, Liping and Jiang, Xue and Liu, Xingzhao and Yan, Junchi and Lyu, Chengqi and Zhang, Wenwei and Chen, Kai},
	title     = {{MMRotate}: A Rotated Object Detection Benchmark Using {PyTorch}},
	booktitle = ACMMM,
	pages     = {7331--7334},
	year      = {2022}
}

@article{CenterMap,
	author  = {Wang, Jinwang and Yang, Wen and Li, Heng-Chao and Zhang, Haijian and Xia, Gui-Song},
	title   = {Learning Center Probability Map for Detecting Objects in Aerial Images},
	journal = TGRS,
	volume  = {59},
	number  = {5},
	pages   = {4307--4323},
	year    = {2021}
}

@article{cheng2022anchor,
	author  = {Cheng, Gong and Wang, Jiabao and Li, Ke and Xie, Xingxing and Lang, Chunbo and Yao, Yanqing and Han, Junwei},
	title   = {Anchor-Free Oriented Proposal Generator for Object Detection},
	journal = TGRS,
	volume  = {60},
	pages   = {1--11},
	year    = {2022}
}

@article{cheng2022dual,
	author  = {Cheng, Gong and Yao, Yanqing and Li, Shengyang and Li, Ke and Xie, Xingxing and Wang, Jiabao and Yao, Xiwen and Han, Junwei},
	title   = {Dual-Aligned Oriented Detector},
	journal = TGRS,
	volume  = {60},
	pages   = {1--11},
	year    = {2022}
}

@article{cheng2023towards,
	author  = {Cheng, Gong and Yuan, Xiang and Yao, Xiwen and Yan, Kebing and Zeng, Qinghua and Xie, Xingxing and Han, Junwei},
	title   = {Towards Large-Scale Small Object Detection: Survey and Benchmarks},
	journal = TPAMI,
	volume  = {45},
	pages   = {13467--13488},
	year    = {2023}
}

@article{han2021align,
	author  = {Han, Jiaming and Ding, Jian and Li, Jie and Xia, Gui-Song},
	title   = {Align Deep Features for Oriented Object Detection},
	journal = TGRS,
	volume  = {60},
	pages   = {1--11},
	year    = {2021}
}

@article{lu2023robust,
	author  = {Lu, Wei and Chen, Si-Bao and Tang, Jin and Ding, Chris H. Q. and Luo, Bin},
	title   = {A Robust Feature Downsampling Module for Remote-Sensing Visual Tasks},
	journal = TGRS,
	volume  = {61},
	pages   = {1--12},
	year    = {2023}
}

@article{LU2026431,
	author  = {Lu, Wei and Li, Hui-Dong and Wang, Chao and Chen, Si-Bao and Ding, Chris H. Q. and Tang, Jin and Luo, Bin},
	title   = {{UnravelNet}: A Backbone for Enhanced Multi-Scale and Low-Quality Feature Extraction in Remote Sensing Object Detection},
	journal = ISPRS,
	volume  = {231},
	pages   = {431--442},
	year    = {2026}
}

@article{nie2022multi,
	author  = {Nie, Guangtao and Huang, Hua},
	title   = {Multi-Oriented Object Detection in Aerial Images with Double Horizontal Rectangles},
	journal = TPAMI,
	volume  = {45},
	number  = {4},
	pages   = {4932--4944},
	year    = {2022}
}

@article{ren2016faster,
	author  = {Ren, Shaoqing and He, Kaiming and Girshick, Ross and Sun, Jian},
	title   = {Faster {R-CNN}: Towards Real-Time Object Detection with Region Proposal Networks},
	journal = TPAMI,
	volume  = {39},
	number  = {6},
	pages   = {1137--1149},
	year    = {2016}
}

@article{Sun2025CTRP,
	author  = {Sun, Peng and Zheng, Yongbin and Xu, Wanying and Li, Jian and Yang, Jiansong},
	title   = {Completing Missing Entities: Exploring Consistency Reasoning for Remote Sensing Object Detection},
	journal = TIP,
	volume  = {35},
	pages   = {569--584},
	year    = {2026}
}

@article{tian2020fcos,
	author  = {Tian, Zhi and Shen, Chunhua and Chen, Hao and He, Tong},
	title   = {{FCOS}: A Simple and Strong Anchor-Free Object Detector},
	journal = TPAMI,
	volume  = {44},
	number  = {4},
	pages   = {1922--1933},
	year    = {2022}
}

@article{xi2024structure,
	author  = {Xi, Yifan and Lu, Ting and Kang, Xudong and Li, Shutao},
	title   = {Structure-Adaptive Oriented Object Detection Network for Remote Sensing Images},
	journal = TGRS,
	volume  = {62},
	pages   = {1--13},
	year    = {2024}
}

@article{xu2020gliding,
	author  = {Xu, Yongchao and Fu, Mingtao and Wang, Qimeng and Wang, Yukang and Chen, Kai and Xia, Gui-Song and Bai, Xiang},
	title   = {Gliding Vertex on the Horizontal Bounding Box for Multi-Oriented Object Detection},
	journal = TPAMI,
	volume  = {43},
	number  = {4},
	pages   = {1452--1459},
	year    = {2020}
}

@article{yang2018automatic,
	author  = {Yang, Xue and Sun, Hao and Fu, Kun and Yang, Jirui and Sun, Xian and Yan, Menglong and Guo, Zhi},
	title   = {Automatic Ship Detection in Remote Sensing Images from {Google Earth} of Complex Scenes Based on Multiscale Rotation Dense Feature Pyramid Networks},
	journal = RS,
	volume  = {10},
	number  = {1},
	pages   = {132},
	year    = {2018}
}

@article{zheng2025spcnet,
	author  = {Zheng, Yunping and Wang, Zejun and Wang, Kunzhi and Jiang, Zhou and Sarem, Mudar},
	title   = {{SPCNet}: Serial Pyramid Convolutional Network for Remote Sensing Object Detection},
	journal = TGRS,
	volume  = {63},
	pages   = {1--10},
	year    = {2025}
}

% Check whether the conference requires a reproducibility checklist to be included in the paper.
% If so, you can uncomment the following line and ajust the path to include it.

\end{document}